\documentclass{article}

\usepackage{arxiv}

\usepackage[utf8]{inputenc} 
\usepackage[T1]{fontenc}    
\usepackage{hyperref}       
\usepackage{url}            
\usepackage{booktabs}       
\usepackage{amsfonts}       
\usepackage{nicefrac}       
\usepackage{microtype}      
\usepackage{lipsum}
\usepackage{graphicx}
\usepackage{subfigure}
\usepackage{amsmath, cuted,stmaryrd}
\usepackage{algorithm}
\usepackage[noend]{algpseudocode}
\usepackage[table]{xcolor}
\usepackage{multirow,booktabs}
\usepackage{diagbox}
\usepackage[normalem]{ulem}
\usepackage[toc,page]{appendix}
\usepackage{biblatex}
\title{Travel time prediction for congested freeways\\with a dynamic linear model}

\author{
 Semin Kwak \\
  School of Electrical Engineering\\
  École Polytechnique Fédérale de Lausanne (EPFL)\\
  Lausanne, Switzerland 1015 \\
  \texttt{semin.kwak@epfl.ch} \\
   \And
 Nikolas Geroliminis \\
  School of Civil Engineering\\
  École Polytechnique Fédérale de Lausanne (EPFL)\\
  Lausanne, Switzerland 1015 \\
  \texttt{nikolas.geroliminis@epfl.ch} 
}

\begin{document}
\maketitle
\begin{abstract}
Accurate prediction of travel time is an essential feature to support Intelligent Transportation Systems (ITS). The non-linearity of traffic states, however, makes this prediction a challenging task. Here we propose to use dynamic linear models (DLMs) to approximate the non-linear traffic states. Unlike a static linear regression model, the DLMs assume that their parameters are changing across time. We design a DLM with model parameters defined at each time unit to describe the spatio-temporal characteristics of time-series traffic data. 
Based on our DLM and its model parameters analytically trained using historical data, we suggest an optimal linear predictor in the minimum mean square error (MMSE) sense.
We compare our prediction accuracy of travel time for freeways in California (I210-E and I5-S) under highly congested traffic conditions with those of other methods: the instantaneous travel time, k-nearest neighbor, support vector regression, and artificial neural network. We show significant improvements in the accuracy, especially for short-term prediction.
\end{abstract}


\section{Introduction}
Travel time prediction is one of the essential features to support successful Intelligent Transportation Systems (ITS). An accurate prediction of travel time not only helps travelers to make decisions about their trips but also enables traffic operators to develop successful control strategies.
This necessity has engaged many researchers on the topic of travel time forecasting despite the vast amount of existing literature. 

The methods for predicting travel time can be categorized into model-based and data-driven approaches \cite{10.1007/978-3-030-05755-8_7}. The model-based methods predict future traffic parameters (e.g., occupancy, flow, or speed) by building a traffic model, such as the Cell Transmission Model \cite{ben2001network, wan2014prediction}, the queuing theory \cite{takaba1991estimation, ben2001network, skabardonis2005real}, or macroscopic traffic flow model \cite{nanthawichit2003application}. These model-based methods provide a straightforward interpretation of the predicted results because of their physical intuition, such as flow dynamics.  

The data-driven methods, on the other hand, predict travel time by extracting specific features from traffic data. Common data-driven methods include Linear Regression \cite{rice2001simple, Zhang:2003bn}, Autoregressive models \cite{yang2009reliability,xia2011multistep, zhang2014component, salamanis2015managing}, Kalman Filtering \cite{Chien:2003gg,kuchipudi2003development,van2008online, achar2019bus} and Bayesian inference~\cite{van2009bayesian,fei2011bayesian}. These methods predict travel time by assuming that all the data satisfies a certain probabilistic distribution. 

Furthermore, the increased accessibility to traffic data and the improved computing power in these days allow researchers to develop more sophisticated data-driven algorithms, such as Support vector regression (SVR) \cite{wu2003travel, castro2009online, gao2016travel}, Artificial neural networks (ANN) \cite{Park:1999ik,dia2001object,dharia2003neural, innamaa2005short,van2005accurate,liu2006predicting,li2014data, hou2018network}, Long Short-Term Memory Network \cite{duan2016travel,liu2017short} and Ensemble learning \cite{hamner2010predicting,zhang2015gradient,petersen2019multi}.

Conversely, the travel time predictors can also be categorized into {\it{direct}} and {\it{indirect}} methods. Direct methods contain a straightforward approach to minimize the error in predicted travel time \cite{rice2001simple,Zhang:2003bn,yang2009reliability,xia2011multistep,zhang2014component,salamanis2015managing,kuchipudi2003development, van2008online, achar2019bus,van2009bayesian,fei2011bayesian,wu2003travel,castro2009online,gao2016travel,Park:1999ik,dia2001object,dharia2003neural,innamaa2005short,hou2018network,duan2016travel,zhang2015gradient}. The main advantage of the direct methods lies in their simplicity since they take into account only the travel time as an output. However, the prediction performance can also be degraded as all the complex traffic characteristics are assumed to be reflected in travel time. Another limitation of these methods is that they require separate models for different circumstances; for example, when the departure time or the origin location change a new model has to be trained for that exact setting. 

In contrast, the indirect methods estimate travel time by predicting future traffics first, such as velocity or occupancy field. Then they use the predicted traffic states to estimate travel time  \cite{Chien:2003gg,hamner2010predicting,robinson2005modeling, Yildirimoglu:2013ip}. By definition, the predicted traffic states can also be re-used to predict those for the next horizon. Contrarily to the direct methods, predicting future traffic parameters allows the model to estimate a travel time for any scheduled departure time or space, which makes the indirect method more versatile.

In this paper, we suggest a method based on a dynamic linear model to predict the velocity field and therefore travel time, which falls into the intersection of the indirect method and the data-driven approach. Using historical data, we analytically find the model parameters in the least-squares sense. We compare the proposed method with four other predictors that are used in the literature: the instantaneous travel time, the k-nearest neighbor \cite{robinson2005modeling}, artificial neural networks \cite{Park:1999ik}, and the support vector regression  \cite{wu2003travel}. Our comparison shows that the proposed method has a great potential to improve the short-term prediction accuracy as well as to become a versatile tool in various traffic situations with this stand-alone model.

\section{Method}
\subsection{Dynamic linear model}

We suggest a dynamic linear model for speed and travel time prediction. The dynamic characteristics allow the model to extract temporal features of the parameters of interest (velocity fields in our case). The model describes a linear relationship between velocities at a specific time $t_k$ and the next step time $t_{k+1}$ using the following equation:

\begin{equation}\label{eqn:piecewiseModel}
{{\bf{v}}_{{k+1}}^d} = {{H}_{k}}{{\bf{v}}_{k}^d} + {{\bf{n}}_{k }^d},\;\forall d,\;\forall k \in \{0,\dots,K-1\}.
\end{equation} 

Here the vector ${\bf{v}}_{k}^{d}$ refers to a velocity vector at time $t_k$ on day $d$, which can be expressed as follows: 
\begin{equation}\label{eqn:velocityVector}
{\bf{v}}_{k}^{d}
=\begin{bmatrix}
v^d\left(x_1,t_k\right)\\
\vdots \\
v^d\left(x_M,t_k\right)
\end{bmatrix}
\in {\mathbb{R}^{M \times 1}},
\end{equation} 
where the constant $M$ represents the number of measured velocities on different locations on a freeway of interest, for example with data from loop detectors. We define a velocity field as a scalar function of time $t$ and position $x$:

\begin{equation}
    v^d\left(x,t\right)\in \mathbb{R},
\end{equation}
where each point of the velocity field represents a measured velocity value. 

In Eq. (\ref{eqn:piecewiseModel}), the second vector ${{\bf{n}}_{k }^d}$ on the right-hand side refers to a noise vector which we assume to follow a Gaussian distribution with a zero mean and a variance ${{{\sigma }}^2}$ under independent and identically distributed (i.i.d.) conditions, i.e.: 
\begin{equation}
{{\bf{n}}_{k }^d} \sim \mathcal{N}\left( {\bf{0}}_M, \sigma^2{I}_M\right),\;\forall k \in \{0,\dots,K-1\}, 
\end{equation}
where ${\bf{0}}_M$ and ${I}_M$ are the vectors with all zero entities and the identity matrix of size $M$, respectively.

Matrix $H_{k}$ in Eq.~(\ref{eqn:piecewiseModel}) is a transition matrix, which represents a linear relationship between the two velocity vectors ${\bf{v}}_{k}^{d}$ and ${\bf{v}}_{k+1}^{d}$ according to the time $t_k$ and $t_{k+1}$. In particular, the diagonal elements of $H_k$ describe a direct temporal relationship at each specific location, whereas the off-diagonal terms of $H_k$ contain the spatio-temporal relationship between two consecutive velocity fields. The first aim of this paper is to find an analytical solution of the transition matrix $H_{k}$ for every possible time $t_k$ so that we build a dynamic transition matrix. 

The model presented in Eq.~(\ref{eqn:piecewiseModel}) suggests three important factors. First, the velocity vector ${\bf{v}}_k^{d}$ is {\it{linearly}} transformed to the vector ${\bf{v}}_{k+1}^{d}$ with an additive Gaussian noise. The linearity and the Gaussian noise assumption allow the transition matrices ${H}_{k}$ for every $k$ to be trained analytically, which will be discussed in the next section. Secondly, the transition matrix is defined at each time unit such that the model captures a non-linear traffic flow over time even though it is based on a linear regression model for a given period. Lastly, the transformation matrix of two consecutive time steps $k$ and $k+1$ is set regardless of different traffic profiles, which means that the matrix ${H}_{k}$ does not depend on, for example, the days of a week. 
We will show later that this framework captures well traffic conditions of different days.

\subsection{Estimation of model parameters}

We shall estimate the transition matrices $H_{k}$ for all $k$ values with historical data set using the least-squares method. Within a set of days $\mathbb{D}$ we choose for estimation (or we call it a day set), Eq.~(\ref{eqn:piecewiseModel}) can be extended as
 \begin{equation}\label{eqn:modelwithmanydata}
{V}_{k + 1}^{\mathbb{D}} = {{H}_{k}}{V}_k^{\mathbb{D}} + {N}_{k}^{\mathbb{D}},
\end{equation} 
where the matrix ${V}_k^{\mathbb{D}}$ is a time-velocity matrix defined for the day set $\mathbb{D}$ for a specific time $t_k$ as a collection of all velocity vectors corresponding to the same time index within $\mathbb{D}$, i.e.:
\begin{equation}\label{eqn:matrixmodel}
{V}_{k}^{\mathbb{D}}  = 
\begin{bmatrix}
{{\bf{v}}^{d_1}_{k}} & {{\bf{v}}^{d_2}_{k}} & \hdots & {{\bf{v}}^{d_{\left|\mathbb{D}\right|}}_{k}}
\end{bmatrix} 
\in {\mathbb{R}^{M \times {\left|\mathbb{D}\right|}}},\;\forall d_i\in\mathbb{D}.
\end{equation} 
Here, the operator $\left|\cdot\right|$ of a set represents the cardinality (the number of elements) of the set. Therefore, the number of the rows and columns represents the data dimension and the size of a day set, respectively.

From Eq. (\ref{eqn:modelwithmanydata}), we shall estimate the transition matrix using the least-squares method, which is also equivalent to the solution of the maximum likelihood method since we assume i.i.d. Gaussian noise~\cite{kailath2000linear}. Therefore, the optimization problem can be stated as:
 \begin{equation}\label{eqn:lsModel}
\mathop {{\mathop{\rm minimize}\nolimits} }\limits_{{{H}_{k}}} \left\| { {{{V}_{k + 1}^{\mathbb{D}}} - {{H}_{k}}{{V}_k^{\mathbb{D}}}} } \right\|_F^2,
\end{equation} 
where the operator ${\left\| {A} \right\|_F} = \sqrt {\text{tr} \left( {{A}{{A}^\top}} \right)}$ and ${\text{tr} \left( {{A}{{A}^\top}} \right)}$ indicates a sum of the all diagonal elements of a matrix ${{A}{{A}^\top}}$.

In order to prevent an ill-posed problem and to give priority to more recent data for better prediction, we introduce an adaptive matrix regularization term with a regularization parameter $\rho$ and a forgetting factor $\lambda$, which is recursively multiplied to old data set, to Eq.~(\ref{eqn:lsModel}) as follows:
\begin{equation}\label{eqn:modifiedOP}
\mathop {{\mathop{\rm minimize}\nolimits} }\limits_{{{H}_{k}}}\; \rho {\lambda ^{\left| {\mathbb{D}} \right|}}\left\| {{{H}_{k}}} \right\|_F^2 + \left\| {\left( {{V}_{k+1}^{\mathbb{D}} - {{H}_{k}}{V}_k^{\mathbb{D}}} \right)}{\Lambda} _{\left| {\mathbb{D}} \right|}^{\frac{1}{2}} \right\|_F^2,
\end{equation} 
where the diagonal matrix ${{\Lambda} _N}$ is defined as follows with the forgetting factor $\lambda$:
 \begin{equation}
{{\Lambda} _N} = 
\begin{bmatrix}
{{\lambda ^{N - 1}}}&0& \cdots &0\\
0&{{\lambda ^{N - 2}}}& \ddots & \vdots \\
 \vdots & \ddots & \ddots &0\\
0& \cdots &0&1
\end{bmatrix}.
\end{equation} 

The first term in Eq.~(\ref{eqn:modifiedOP}) is the regularization term; its major role is to prevent the transition matrices from overfitting to a small training data set. The term also allows reliable estimation of the transition matrix numerically, which is described in Appendix \ref{app:regul}. 

The forgetting factor, on the other hand, decreases the weight of old data exponentially during the recursive training process. For instance, when $\lambda=0.995$, a set of data a year ago is penalized by the factor of $(0.995)^{365}=0.16$. This forgetting factor also allows the regularization term to vanish, adapting to the size of the training set since the term converges to zero when the number of elements in $\mathbb{D}$ is getting bigger.
The modified problem in Eq.~(\ref{eqn:modifiedOP}) will be equivalent to the original problem of Eq.~(\ref{eqn:lsModel}) when we set $\rho=0$ and $\lambda=1$, which means no regularization and no forgetting process.

The optimization problem in Eq.~(\ref{eqn:modifiedOP}) can be analytically solved, and we derive it in Appendix \ref{app:RLSsolution}. Here we present the solution:
 \begin{equation} \label{eqn:modifiedOPSolution}
{{{\bar H} }^{\mathbb{D}}}_{k} = {V}_{k+1}^{\mathbb{D}}{{\Lambda} _{\left|\mathbb{D}\right|}}{\left( {{V}_k^{\mathbb{D}}} \right)^\top}{\left( {{V}_{t }^{\mathbb{D}}{{\Lambda} _{{\left|\mathbb{D}\right|}}}{{\left( {{V}_k^{\mathbb{D}}} \right)}^\top} + \rho {\lambda ^{{\left|\mathbb{D}\right|}}}{{I}_M}} \right)^{ - 1}},
\end{equation} 
where the notation ${\bar A}$ represents an estimator of $A$. 

One popular property of the solutions of the least squares problem is that they can be updated with new observations~\cite{kailath2000linear}. The updating method not only prevents increasing memory size, but also makes computation time consistent since the procedure only needs a pre-trained model and a new observation for an update. This property can be essential to support accurate travel time prediction because the system should be up-to-date with time. We describe the implementation of an updating algorithm for Eq.~(\ref{eqn:modifiedOPSolution}) in Appendix~\ref{app:recursive}.

\begin{figure}[t!]
\centering
\includegraphics[width=0.6\columnwidth]{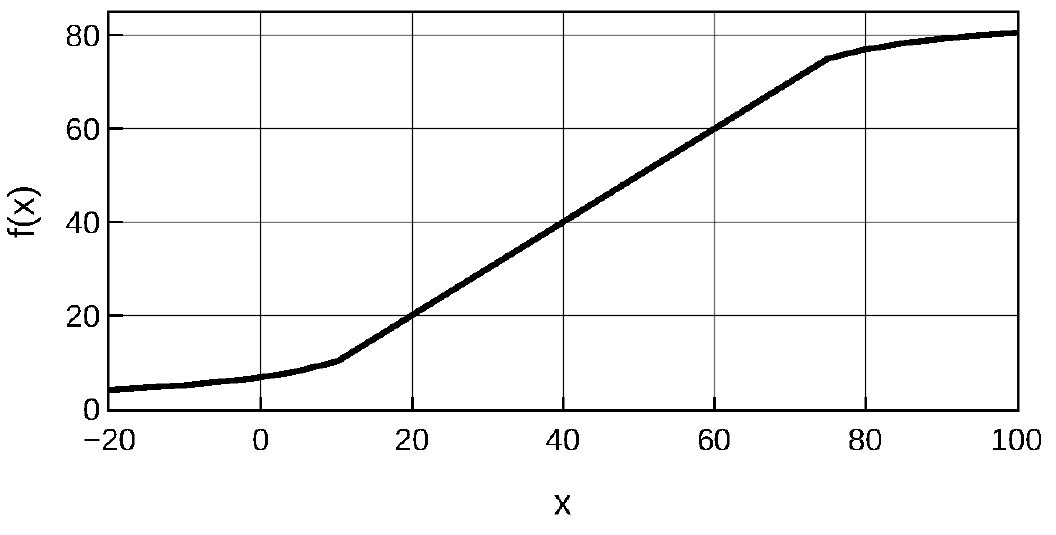}
\caption{Post-processing function for the freeways studied in this paper. The function makes the output values fall within a reasonable speed range, i.e., $0 \leq f(x) \leq 85$.}
\label{fig:postProcess}
\end{figure}

\subsection{Velocity prediction}\label{section:prediction}
With the transition matrices ${{{\bar H} }^{\mathbb{D}}}_{k}$ introduced in the previous section, we can now predict the velocity vectors for traffic forecasting, which is the second aim of this paper. We start by setting a notation of a predictor for $i$-step ahead at time step $k$: 
\begin{equation}
{{{\bf{v}}_{k+i|k}^{\tilde d}}}\text{ for }i = 1,2, \cdots\text{ and }\tilde d \notin \mathbb{D}.   
\end{equation}

Assuming that the trained transition matrix is close enough to the truth i.e., ${\bar H}_{k}^{\mathbb{D}} \approx {{H}_{k}}$ for all $k$, the velocity vector ${{{\bf{v}}_{k+i}^{\tilde d}}}$ is written as follows using Eq.~(\ref{eqn:piecewiseModel}):

\begin{align}
{{\bf{v}}_{k+i}^{\tilde d}} &= {{\bar H}_{k+i - 1}^{\mathbb{D}}}{{\bf{v}}_{k+i - 1}^{\tilde d}} + {{\bf{n}}_{k+i-1}^{\tilde d}}= {{\bar H}_{k+i - 1}^{\mathbb{D}}}\left( {{{{\bar H}}_{k+i - 2}^{\mathbb{D}}}{{\bf{v}}_{k+i - 2}^{\tilde d}} + {{\bf{n}}_{k+i - 2}^{\tilde d}}} \right) + {{\bf{n}}_{k+i-1}^{\tilde d}}\\
&\quad\vdots \nonumber\\
&= {\bar H}_{k+i - 1\shortleftarrow k}^{\mathbb{D}}{{\bf{v}}_k^{\tilde d}} +{{\bf{n}}_{k+i-1\shortleftarrow k}^{\tilde d}},\label{eqn:derivation_predictor}
\end{align}
where
\begin{equation}\label{eqn:h_propagation}
    {\bar H}_{k+i - 1\shortleftarrow k}^{\mathbb{D}}=\prod\limits_{j= 1}^i {{\bar H}_{k+i-j}^{\mathbb{D}}},
\end{equation}
\begin{equation}\label{eqn:n_propagation}
    {{\bf{n}}_{k+i-1\shortleftarrow k}^{\tilde d}}=\sum\limits_{j= 1}^{i - 1} {{\bar H}_{k + i-j}^{\mathbb{D}} {{\bf{n}}_{k+i-j-1}^{\tilde d}}}+{{\bf{n}}_{k+i-1}^{\tilde d}}.
\end{equation}
The noise vector in Eq.~(\ref{eqn:derivation_predictor})  follows a zero mean Gaussain vector with a covariance ${\Sigma}$ since a linear combination of zero mean Gaussian random variable follows another zero mean Gaussian random variable~\cite{feller2008introduction}. As a result, 
\begin{equation}
{{\bf{v}}_{k+i}^{\tilde d}} \sim \mathcal{N}\left( {\bar H}_{k+i - 1\shortleftarrow k}^{\mathbb{D}}{{\bf{v}}_k^{\tilde d}},\Sigma \right).    
\end{equation}

We choose a predictor as the maximizer of the above density function:
\begin{equation}\label{eqn:generalizedPrediction}
{{\bf{v}}_{k+i|k}^{\tilde d}} = {\bar H}_{k+i - 1\shortleftarrow k}^{\mathbb{D}}{{\bf{v}}_k^{\tilde d}}.
\end{equation}

This predictor is also an optimal estimator of the linear minimum mean square error (LMMSE)~(Appendix \ref{app:lmmse}). 
Therefore, Eq.~(\ref{eqn:generalizedPrediction}) shows that the best linear predictor ${{\bf{v}}_{k+i|k}^{\tilde d}}$ is the propagation of the current measurement ${\bf{v}}_k^{\tilde d}$ through the trained transition matrices from ${{\bar H}_{k}^{\mathbb{D}}}$ to ${{\bar H}_{k+i - 1}^{\mathbb{D}}}$.

\begin{figure*}[!t]
   \centering
   \subfigure[Freeway I5-S]{\frame{\includegraphics[width=0.45\textwidth]{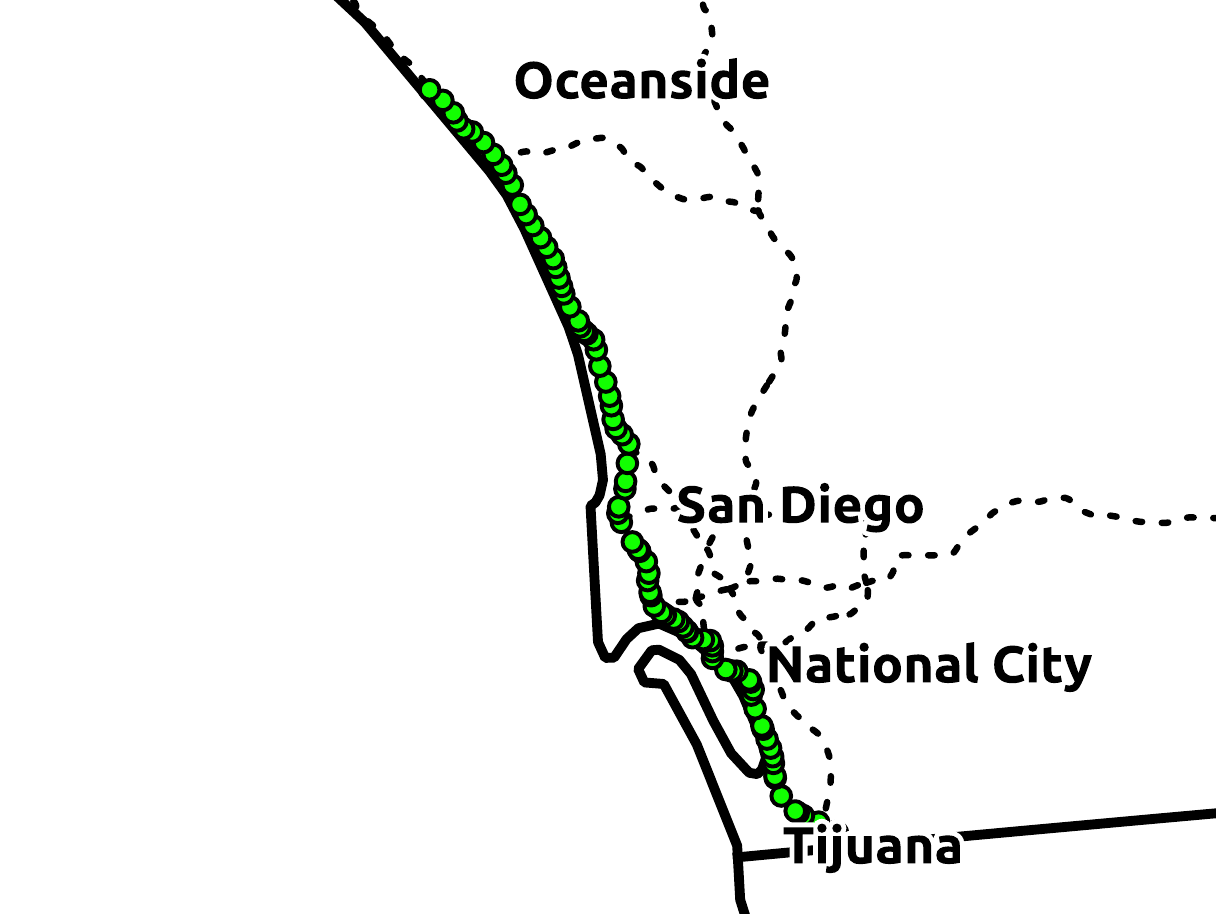}}}\hfill
   \subfigure[Freeway I210-E]{\frame{\includegraphics[width=0.45\textwidth]{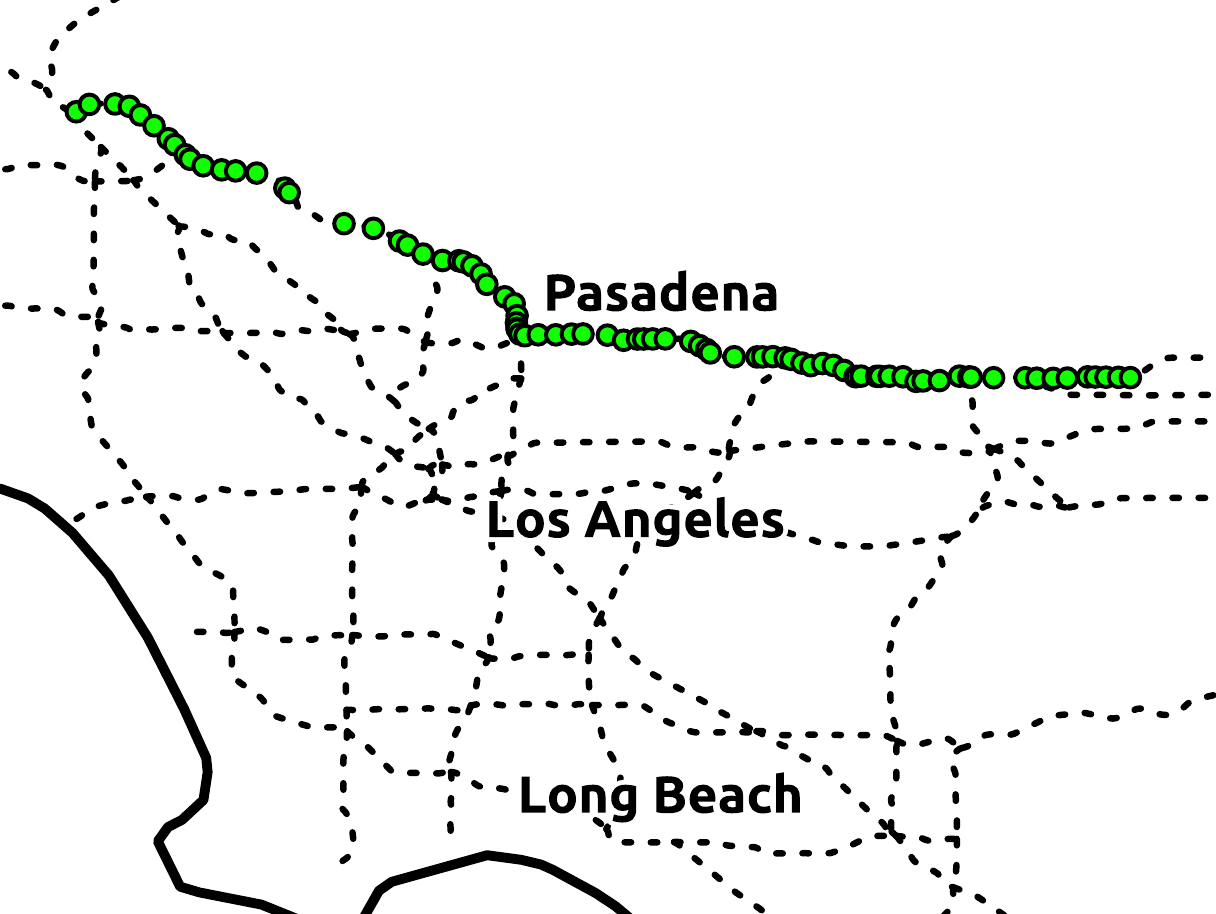}}}
   \caption{Detector locations on the freeways I5-S and I210-E. The 88 loop detectors along the freeway I5-S and the 83 loop detectors along the freeway I210-E are used in this paper (green dots on both figures). In the names of the freeways, S (south) and E (east) represent the direction of the freeways.}
   \label{fig:exp_area}
\end{figure*}

However, in rare cases, an unbounded solution ${{\bf{v}}_{k+i|k}^{\tilde d}}$ can have a negative speed or an unrealistically high speed due to the Gaussian noise assumption. 
This kind of wrong estimations severely distort the calculation of travel time. In order to correct this effect, we design a post-processing function $f\left(x\right)$:
\begin{equation}\label{eqn:postProcess}
f\left( x \right) = \left\{ {\begin{array}{*{20}{c}}
{b \cdot \frac{ {a\left( {x - {\tau _l}} \right)} }{{1 + \left| {a\left( {x - {\tau _l}} \right)} \right|}} + {\tau _l}}&{x < {\tau _l}}\\
x&{{\tau _l} \le x \le {\tau _u}}\\
{b \cdot \frac{ {a\left( {x - {\tau _u}} \right)} }{{1 + \left| {a\left( {x - {\tau _u}} \right)} \right|}} + {\tau _u}}&{x > {\tau _u}}
\end{array}} \right.,
\end{equation} 
where the constants $a$ and $b$ are smoothing parameters, and $\tau_l$ and $\tau_u$ are threshold parameters. We empirically set the smoothing parameters $a$ and $b$ to be $0.05$ and $10$, respectively. We also have empirically chosen the lower threshold value $\tau_l$ and the upper threshold value $\tau_u$ as $10$ and $75$ (mph), respectively.

Figure~\ref{fig:postProcess} shows the post-processing function with the chosen parameter sets. The input $x$ in this example is a velocity value. When $x$ is below the lower threshold of $10$, the function deflects the values to be always positive. When $x$ is above the upper threshold of $75$, the function makes the output converging to the upper limit, which is $85$ miles per hour in our case. Between the two boundaries, it does not change the input value. We have tested different sets of the smoothing and threshold parameters and found that it has little impact on prediction results. 

We apply this post-processing function in Eq.~(\ref{eqn:postProcess}) to Eq.~(\ref{eqn:generalizedPrediction}) recursively at each step of multiplication so that we can exclude the invalid estimations. The following shows the detailed procedure.

\begin{equation}\label{eqn:prediction_pp}
\begin{aligned}
    {{\bf{v}}}_{k+1|k}&=f\left(\bar{H}_{k}{\bf{v}}_k\right)\\
    {{\bf{v}}}_{k+2|k}&=f\left(\bar{H}_{k+1}{{\bf{v}}}_{k+1|k}\right)\\
    &\vdots \\
    {{\bf{v}}}_{k+i|k}&=f\left(\bar{H}_{k+i-1}{{\bf{v}}}_{k+i-1|k}\right)
\end{aligned}
\end{equation}

\subsection{Travel time estimation}\label{section:ett}
For estimating the travel time of a moving vehicle, we assume that the vehicle experiences a velocity field, which is a function of time $t$ and space $x$, and we know the exact continuous velocity field $v\left(t,x\right)$. Then we can calculate the increment of time $\Delta t$ after traveling a distance $\Delta x$ as
\begin{equation}
    \Delta t=\frac{1}{v\left(t,x\right)}\Delta x,
\end{equation}
since $v=dx/dt$. Consequently, a travel time at time $t_0$ given a velocity field $v\left(t,x\right)$ is computed recursively as follows:
\begin{algorithm}[H]
\caption{Numerical calculation of travel time}\label{alg:ett}
\begin{algorithmic}[1]
\renewcommand{\algorithmicrequire}{\textbf{Input:}}
\renewcommand{\algorithmicensure}{\textbf{Output:}}
\Require the velocity field $v\left(t,x\right)$; the departure time and location, $t_0$ and $x_0$; the location of the destination $x_M$; and the space increment $\Delta x$
\Ensure the travel time
\State \textbf{Initialization:}
$t \gets t_0$, $x \gets x_0$
\While {$x<x_M$}
\State $t\gets t+\frac{1}{v\left(t,x\right)}\Delta x$
\State $x\gets x+\Delta x$
\EndWhile
\Return $t-t_0$
\end{algorithmic}
\end{algorithm}

In reality, we only know a discretized velocity field instead of a continuous one. For our study, we know velocities at each sensor (every 0.7 miles on average) every 5 minutes. We generate the continuous velocity field by interpolating the discretized velocity field with linear bivariate B-spline curve fitting.

\subsection{Performance measures for comparison with other methods}

To measure the performance of our prediction, we use the absolute percentage error (APE) and the mean absolute percentage error (MAPE), which are defined as:
\begin{equation}
{\text{APE}}\left(t\right)=100\cdot{\left| \frac{a\left(t\right)-p\left(t\right)}{a\left(t\right)} \right|,}
\end{equation}
\begin{equation}
{\text{MAPE}}\left( \mathbb{T} \right) = \frac{{1}}{{\left| \mathbb{T} \right|}}\sum\limits_{t \in \mathbb{T}} {\text{APE}}\left(t\right),
\end{equation}
where the set $\mathbb{T}$ represents a set of time elements to examine. The values $a\left(t\right)$ and $p\left(t\right)$ are respectively the actual travel time and the predicted travel time when departed at time $t$. The MAPE estimates the mean deviation of estimation to the ground truth (i.e., the experienced travel time) in percentage (\%) unit.

\section{Results and Discussions}
In this section, we employ the proposed method to predict traffic flow and thus travel time using real-world data. We examine the performance of the proposed method by comparing predicted travel time with that of other existing predictors.

\begin{figure*}[t!]
   \centering
   \subfigure[Freeway I5-S (weekdays)]{\includegraphics[width=0.45\textwidth]{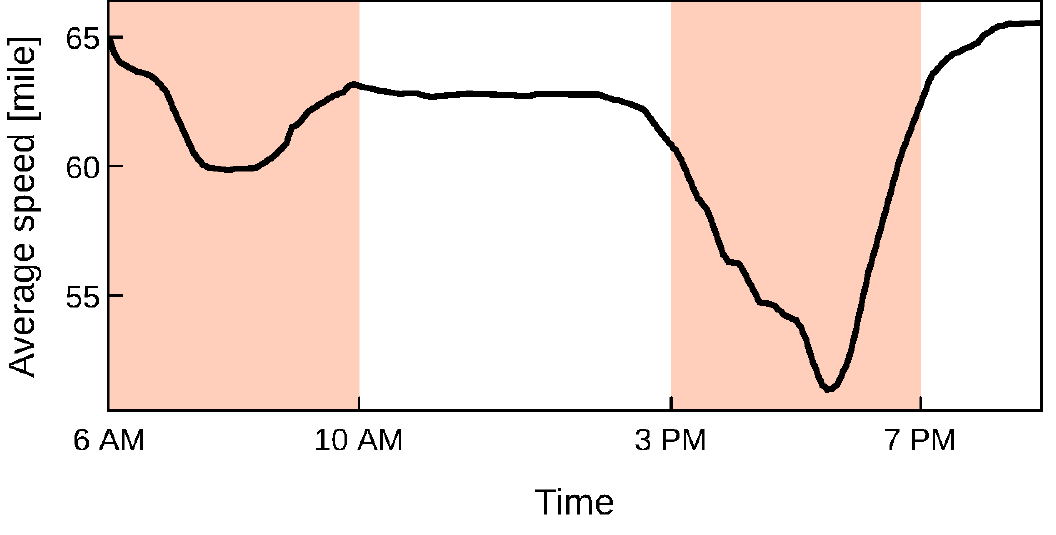}}\hfill
   \subfigure[Freeway I210-E (weekdays and Saturday)]{\includegraphics[width=0.45\textwidth]{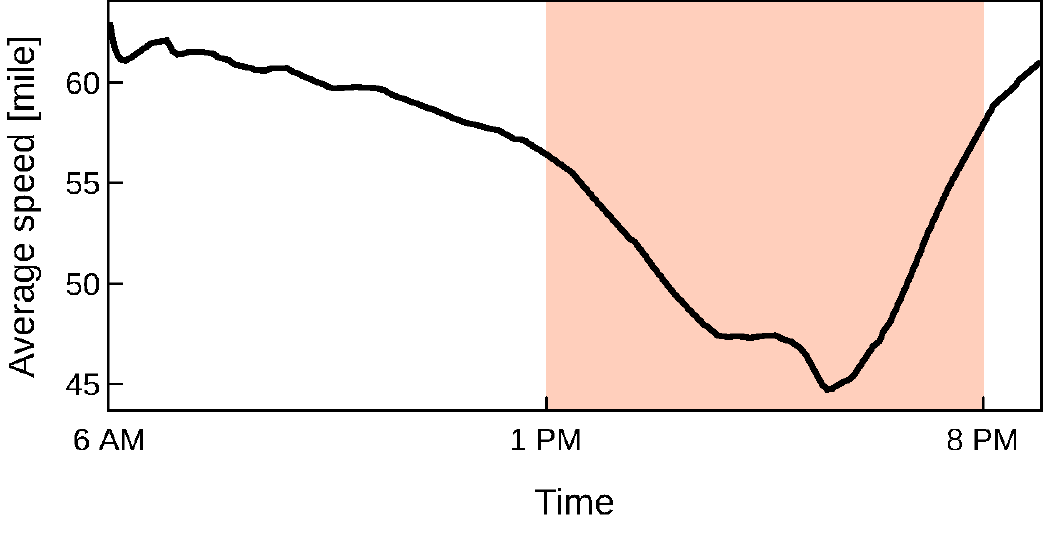}}
   \caption{Average speed by time of test set for each freeway. The peak periods are also defined as the area with color.}
   \label{fig:peakPeriod}
\end{figure*}

\subsection{Traffic data}
We use traffic data of two different freeways in California having different traffic profiles: Freeway I5-S and Freeway I210-E\footnote{The dataset is available: https://doi.org/10.5281/zenodo.3479437}.
Along the two freeways, there are respectively $88$ and $83$ loop detectors within the area of our examination (Fig.~\ref{fig:exp_area}). The total length of the corridor along I5-S is $58.33$ miles, and that of I210-E is $52.14$ miles. The loop detectors collect measurements (flow and occupancy data) every $30$ second, and we use $5$ minutes aggregated speed data, which is processed by the Caltrans Performance Measurement System (PeMS).

From PeMS, we extracted one-year traffic data of both freeways (2012 for I5-S and 2015 for I210-E) for experiments. We allocated the first 70\% of traffic data (from January 1st to September 12th) as a training set, the next 15\% of data as a validation set (from September 13th to November 11th), and the last 15\% of data as a test set (from November 12th to December 31st) for all the experiments.

We have considered the traffic data from 6 AM to 9 PM only and divided the data into two groups: a peak period and an off-peak period (Fig.~\ref{fig:peakPeriod}). Since the two freeways have very different traffic profiles, we have defined the peak and off-peak periods differently for each freeway. For Freeway I5-S, the peak period is defined as 6 - 10 AM (morning peak) and 3 - 7 PM (evening peak) on weekdays (from Mondays to Fridays); for Freeway I210-E, it is defined as 1 - 8 PM (afternoon peak) every day except Sundays. The off-peak periods are defined as a complementary set of the corresponding peak periods. Figure~\ref{fig:peakPeriod} illustrates them straightforwardly.

\begin{table}
\caption{Mean absolute percentage error (MAPE) on the validation sets for Freeways I5-S and I210-E with different hyper-parameter pairs}\label{tab:hyperparameter}
\centering
\subfigure[Freeway I5-S]{
\begin{tabular}{rrrrrr}
\toprule
{\backslashbox{$\rho$}{$\lambda$}} &  1.000 &  0.999 &  0.995 &  0.990 &  0.950 \\
\midrule
0 &  3.447 &  3.433 &  3.414 &  3.432 &  5.134 \\
1 &  3.441 &  3.428 &  3.411 &  3.430 &  5.134 \\
3 &  3.430 &  3.418 &  3.405 &  3.427 &  5.134 \\
10 &  3.395 &  3.388 &  3.385 &  3.415 &  5.134 \\
30 &  3.318 &  3.319 &  3.340 &  3.385 &  5.134 \\
100 &  3.183 &  3.187 &  3.232 &  3.317 &  5.133 \\
300 &  3.071 &  3.074 &  3.106 &  3.202 &  5.130 \\
1000 &  2.996 &  2.987 &  2.982 &  3.045 &  5.119 \\
3000 &  3.003 &  2.987 &  \textbf{2.936} &  2.937 &  5.091 \\
10000 &  3.244 &  3.192 &  3.043 &  2.926 &  5.003 \\
\bottomrule
\end{tabular}
}
\subfigure[Freeway I210-E]{
\begin{tabular}{rrrrrr}
\toprule
{\backslashbox{$\rho$}{$\lambda$}} &  1.000 &  0.999 &  0.995 &  0.990 &  0.950 \\
\midrule
0     &  4.842 &  4.879 &  5.048 &  5.280 &  7.456 \\
1     &  4.858 &  4.896 &  5.063 &  5.290 &  7.455 \\
3     &  4.848 &  4.888 &  5.058 &  5.288 &  7.454 \\
10    &  4.816 &  4.859 &  5.037 &  5.275 &  7.453 \\
30    &  4.747 &  4.793 &  4.986 &  5.239 &  7.453 \\
100   &  4.641 &  4.673 &  4.865 &  5.143 &  7.451 \\
300   &  \textbf{4.602} &  4.622 &  4.729 &  4.998 &  7.449 \\
1000  &  4.643 &  4.637 &  4.672 &  4.808 &  7.443 \\
3000  &  4.806 &  4.781 &  4.718 &  4.745 &  7.427 \\
10000 &  5.318 &  5.232 &  4.965 &  4.811 &  7.369 \\
\bottomrule
\end{tabular}
}
\end{table}
\begin{figure*}
   \centering
   \subfigure[Ground truth (I5-S)]{\includegraphics[width=0.45\textwidth]{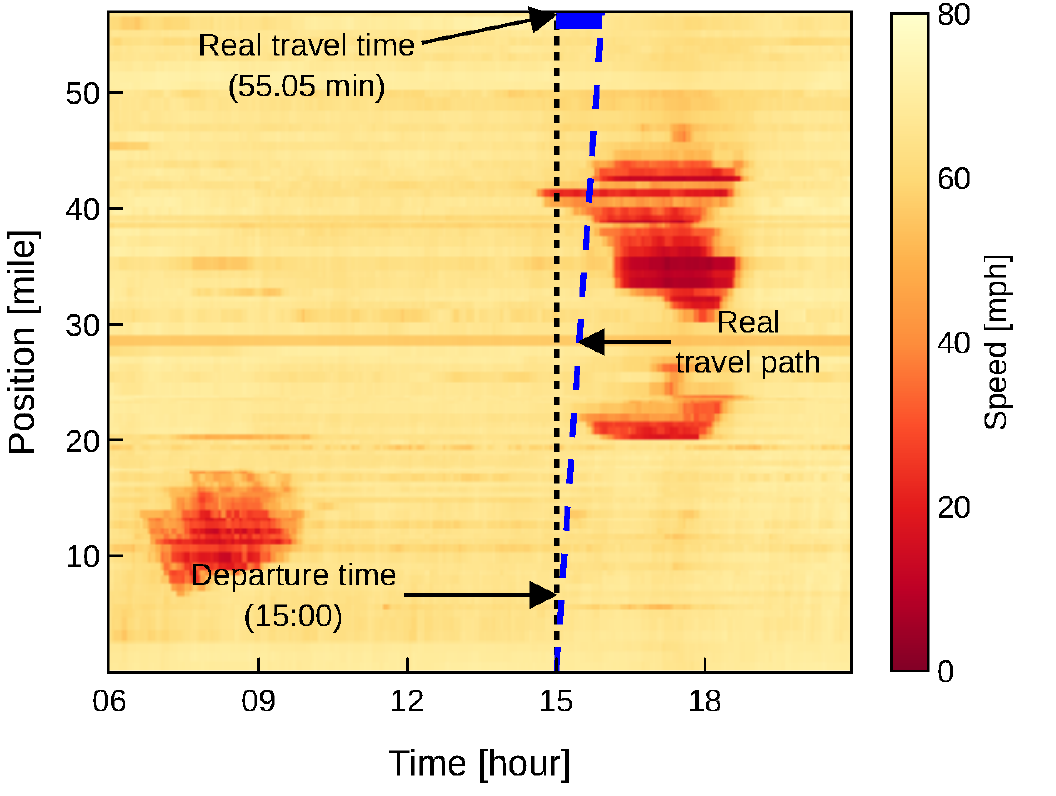}}\hfill
   \subfigure[Ground truth (I210-E)]{\includegraphics[width=0.45\textwidth]{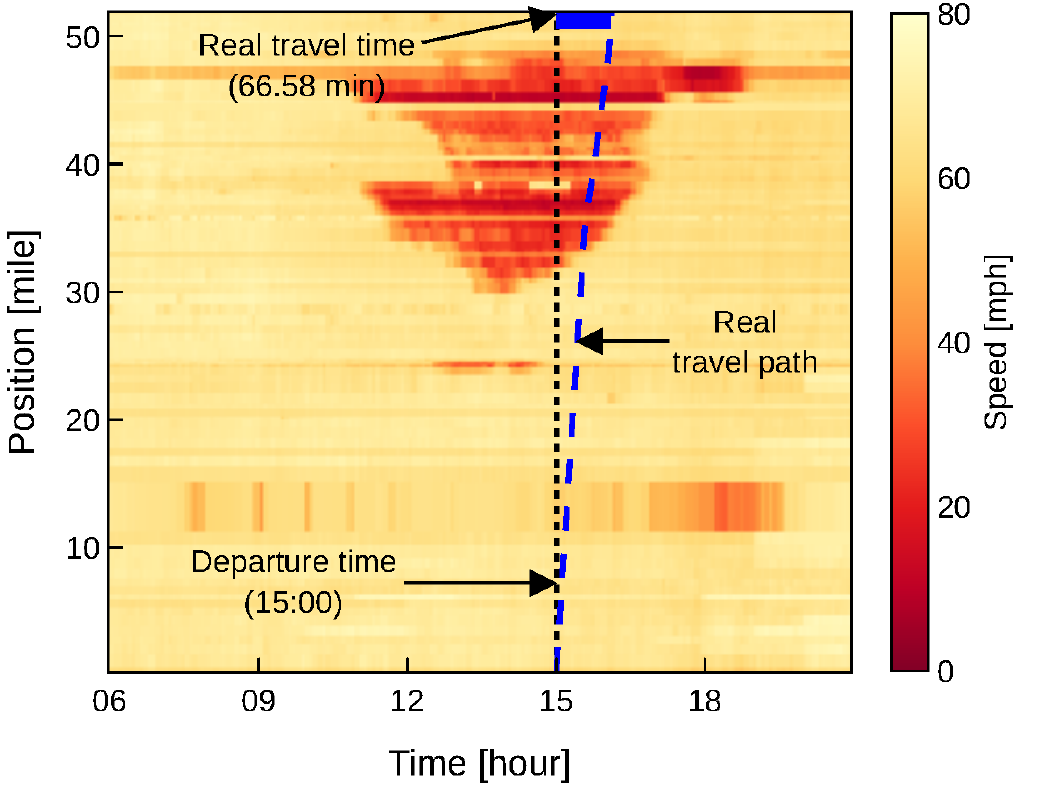}}
   \subfigure[Prediction (I5-S)]{\includegraphics[width=0.45\textwidth]{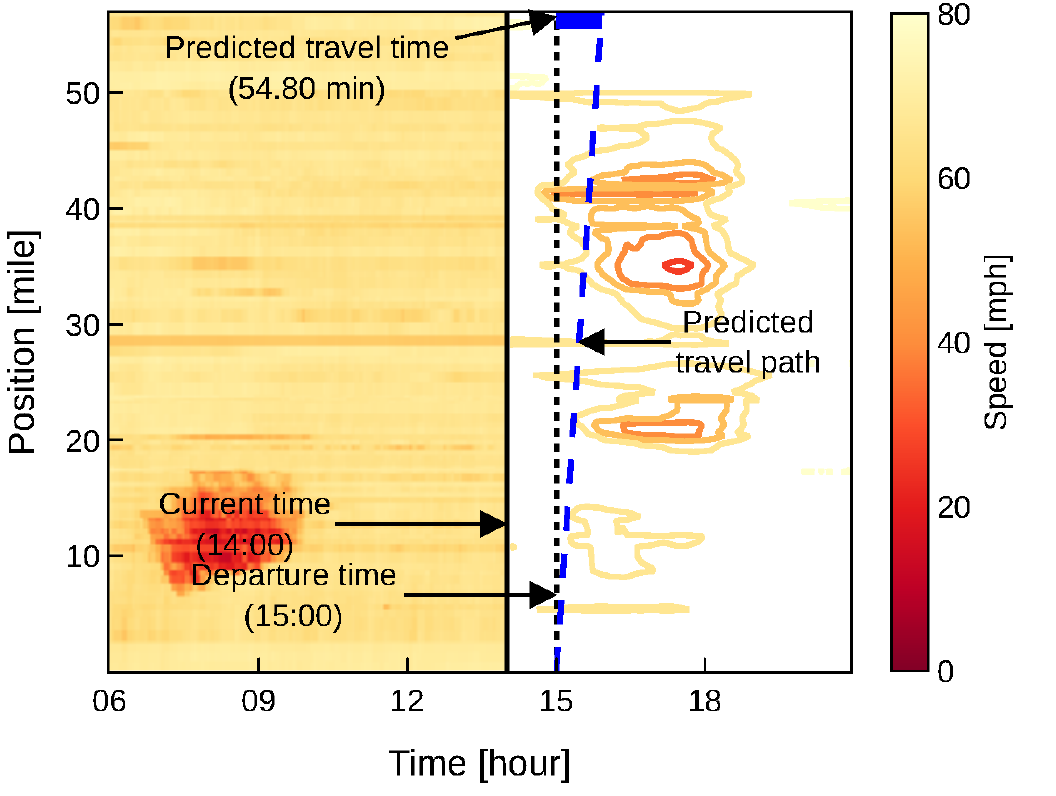}}\hfill
   \subfigure[Prediction (I210-E)]{\includegraphics[width=0.45\textwidth]{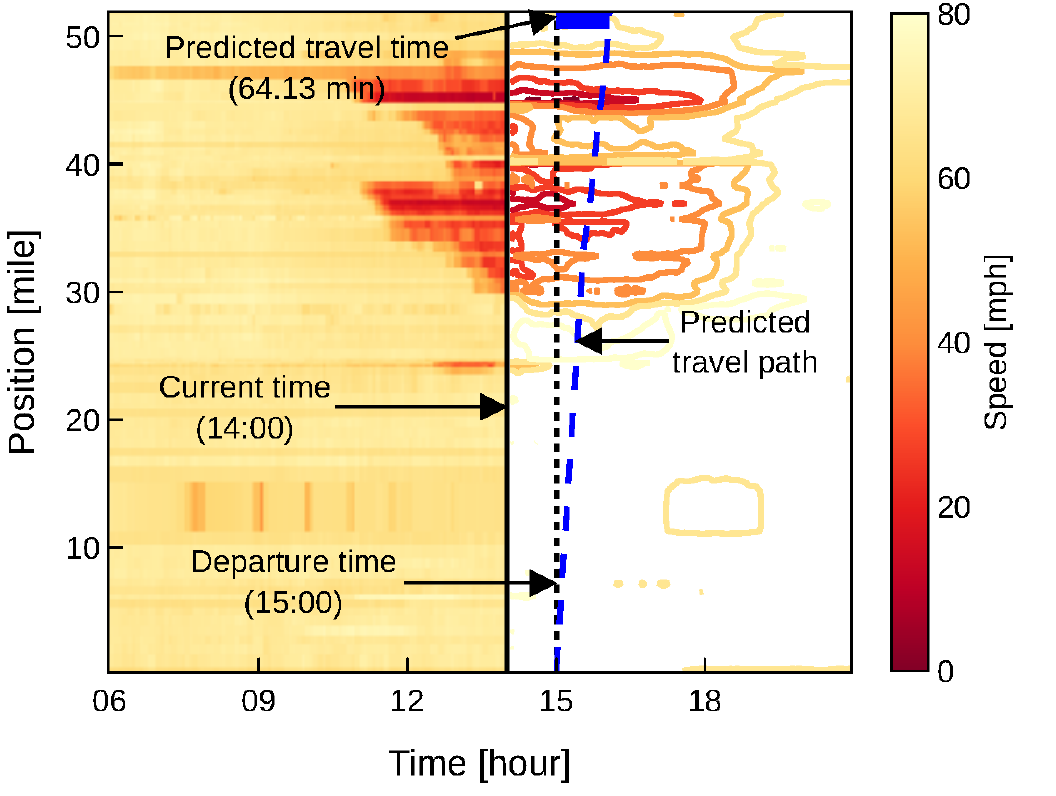}}
   \caption{Velocity field of freeway I5-S on December 4th (Tuesday), 2012 and I210-E on December 25th (Friday), 2015. (a) and (b): The ground truths and (c) and (d): predicted velocity fields at 2 PM and afterwards using the proposed method, which is represented as the contour plots. The blue dashed line represents the travel path of a vehicle at each velocity field. The difference between the departure time and the arrival time is the travel time, which is marked as a blue line on the upper horizontal axis.}
   \label{fig:predictedSpeedProfile}
\end{figure*}

\subsection{Determining hyper-parameters}
\begin{figure*}[!h]
   \centering
    \subfigure[Freeway I5-S: Peak period]{\includegraphics[width=0.47\textwidth]{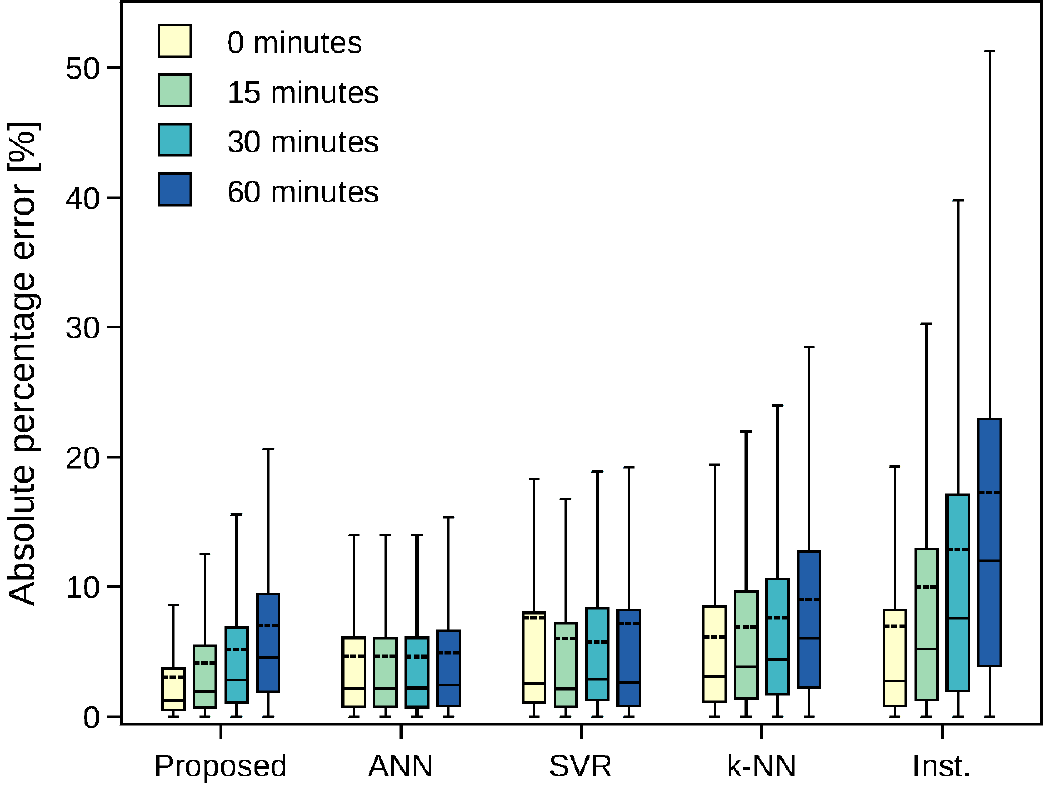}}\hfill
   \subfigure[Freeway I210-E: Peak period]{\includegraphics[width=0.47\textwidth]{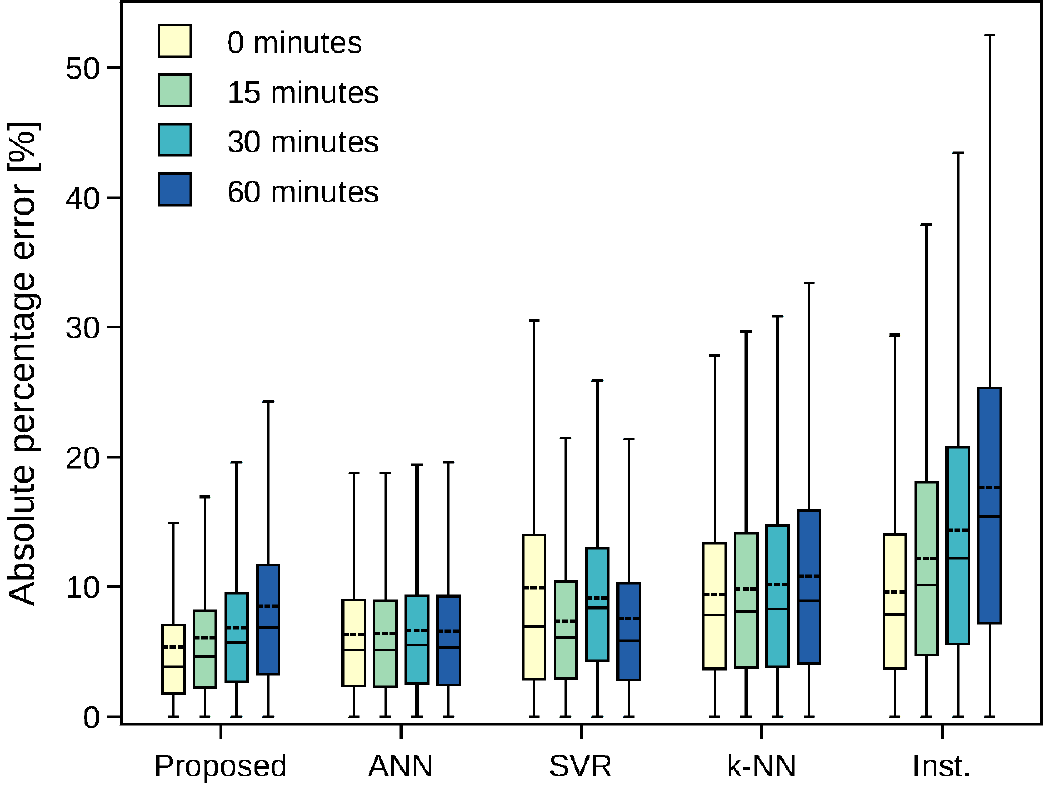}}\\
   \subfigure[Freeway I5-S: Off-peak period]{\includegraphics[width=0.47\textwidth]{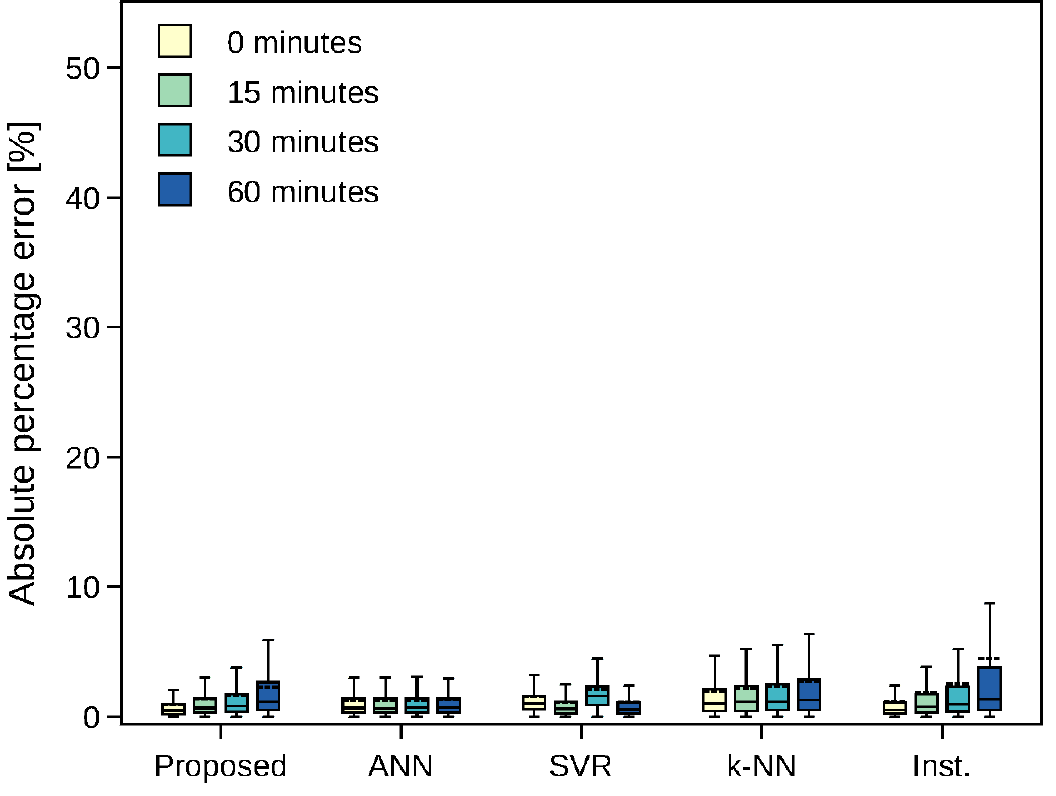}}\hfill
   \subfigure[Freeway I210-E: Off-peak period]{\includegraphics[width=0.47\textwidth]{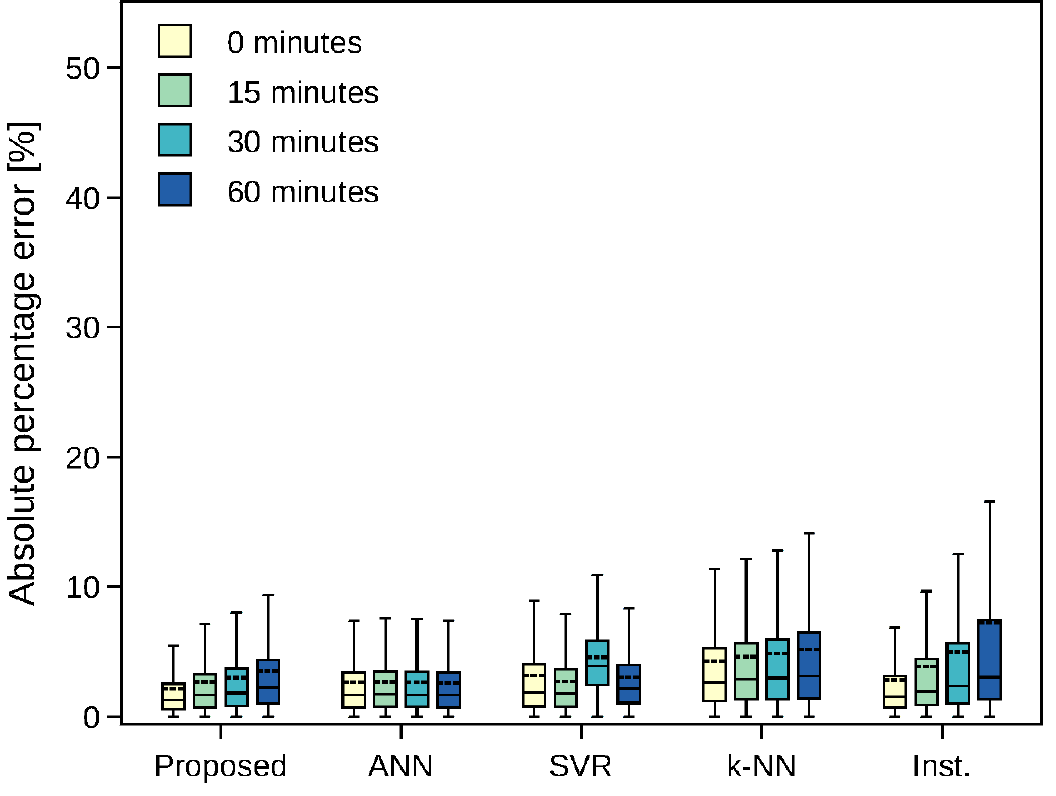}}
   \caption{Absolute percentage errors (APE) of 5 different travel time forecasters with various horizons. Two examples of freeways in California, I5-S and I210-E, are studied during their peak periods (a) and (b); and their off-peak periods (c) and (d). Inside the box plots, the medians and mean values are marked as solid and dashed bars, respectively; different colors represent different prediction horizons.}
   \label{fig:prediction_horizon}
\end{figure*}
Using Eq.~(\ref{eqn:modifiedOPSolution}), we have trained transition matrices with different pairs of the regularization parameter $\rho$ and the forgetting factor $\lambda$. For each freeway, we have trained the transition matrix by all possible combinations of the following sets:
\begin{equation}
\begin{aligned}
\rho&\in \{0, 0.1, 0.3, 1, 3, 10, 30, 100, 300, 1000, 3000, 10000\},\\
\lambda&\in \{1, 0.999, 0.995, 0.99, 0.95\}.
\end{aligned}
\end{equation}

Table~\ref{tab:hyperparameter} shows the MAPE of travel time on the validation sets of the two freeways (peak periods only) by varying the hyper-parameters. The MAPE is not very sensitive to the regularization parameters, but it is influenced by the forgetting factors, as when the value of the forgetting factor is too low, this leads to a training of the transition matrices with not enough data. For each case of the freeways, we have chosen the optimal pair among the tested parameter sets, which are:
\begin{equation}
    \left(\rho,\lambda\right)=
    \left\{ {
    \begin{matrix}
    \left(3000,0.995\right)&\text{for I5-S}\\
    \left(300,1\right)&\text{for I210-E}
    \end{matrix}
    } \right..
\end{equation}

We show that the optimal pairs of hyper-parameters chosen above work well for traffic prediction by showing an example of predicted velocity fields (Fig.~\ref{fig:predictedSpeedProfile}). The prediction results (the contour plot) in Fig.~\ref{fig:predictedSpeedProfile} (c) and (d) show similar patterns to the ground truths (Fig.~\ref{fig:predictedSpeedProfile} (a) and (b)), which confirms that the chosen hyper-parameters are functioning well for predicting speeds and travel time.

\subsection{Comparison of travel time with different forecasters}

We examine the performance of our proposed method by comparing its performance with that of different prediction methods. We have chosen four various forecasters: the instantaneous travel time forecaster (abbreviated to \textit{inst.}) as a real-time measurement-based method; the k-nearest neighbor (k-NN) as a historical data-based method; the support vector regression (SVR) and the vanilla artificial neural network (ANN) as representatives for direct methods. All the details of implementing these methods are explained in Appendix~\ref{app:methods}.

Specifically, we evaluate travel time using these methods with different prediction horizons. We define a travel time at time $t$ with a prediction horizon $h$-minutes as a travel time that a vehicle will experience when it departs $h$-minutes after the time $t$. For example, Fig.~\ref{fig:predictedSpeedProfile} (c) and (d) show travel time prediction with a $60$-minute horizon at the current time of 2 PM. We assign four different values for $h$: 0, 15, 30, and 60 minutes.

Figure~\ref{fig:prediction_horizon} shows the average prediction errors (APE) of the results on the test sets. It shows that the proposed method always gives the best accuracy among others when $h = 0$ minute, for both freeways and in both peak and off-peak periods. For longer horizons, the performance of the proposed method is comparable to that of ANN and SVR, whereas it always performs better than k-NN and \textit{inst.} in these results.

First of all, it is surprising that the proposed method has comparable errors with that of ANN and SVR for longer horizons. The ANN and SVR are direct methods, which means that they have a separate model for each horizon and each one has been trained independently. The proposed method, on the other hand, is an indirect method, which predicts the travel time of longer horizons based on previous predictions. In other words, it uses a model trained only once for all the horizons. 

One could understand this from its superior performance at the 0-minute horizon. As it is seen in all the sub-figures of Fig.~\ref{fig:prediction_horizon}, the proposed method starts from a very small error, and then the error starts to increase gradually when extending the prediction horizon. This is simply due to the aggregate noise in  Eq.~(\ref{eqn:derivation_predictor}). From the fact that covariance of the sum of two Gaussian random variables is always greater than the variance of each variable, the sum of the noise terms in Eq.~(\ref{eqn:n_propagation}) always produces larger covariance and therefore more substantial errors. However, since its initial 0-minute horizon error is very small compared to the other methods, the errors can remain relatively small even when the noise propagates and accumulates with time. 

Compared to the other indirect methods, which are k-NN and instantaneous travel time forecaster, the proposed method shows better prediction regardless of traffic profiles and prediction horizons. We can find the reason by looking into the type of data that are considered in each method. The k-NN is highly dependent on historical data, whereas the instantaneous travel time forecaster uses only the real-time traffic measurement. Our prediction algorithm (Eq.~(\ref{eqn:prediction_pp})), on the other hand, utilizes both the real-time measurement (${\bf{v}}_k$) and the historical information that is considered in the transition matrix  ($\bar{H}_{k}$). This explains why it outperforms the other two methods. 

Table~\ref{tab:horizon} shows the {\it{improvement rates}}, which indicate how much the accuracy (MAPE) of travel time prediction is improved compared to that of the instantaneous travel time. For instance, in the case of Freeway I5-S during the peak periods, according to Table~\ref{tab:horizon} (a), the proposed method improves the prediction accuracy compared to instantaneous travel time by 56\% with the 0-minutes horizon. In contrast, ANN and SVR improve that by 33\% and -9\%, respectively. 

Table~\ref{tab:horizon} also confirms that the proposed method has the best prediction accuracy among all five forecasters for short horizons ($h = 0, 15$ min) and comparable performances to the best one for longer horizons ($h = 30, 60$ min). This result is promising since our approach has an additional degree of freedom to be used for arbitrary departure time and various starting points. 

\begin{table*}[t!]
\caption{Improvement rate of methods on test sets}\label{tab:horizon}
\centering
\subfigure[Freeway I5-S]{
\begin{tabular}{lcccccccc}
\toprule
& \multicolumn{4}{c}{Prediction horizon (minutes)} \\ & 0 & 15 & 30 & 60  \\
\midrule
Proposed         &             \textbf{0.56} &              \textbf{0.58} &               0.60 &               0.59 \\
ANN              &              0.33 &              0.54 &              \textbf{0.64} &   \textbf{0.71} \\
SVR              &             -0.09 &              0.40 &              0.55 &               0.58 \\
k-NN &              0.12 &              0.31 &               0.41 &               0.48 \\
\bottomrule
\end{tabular}}
\subfigure[Freeway I210-E]{
\begin{tabular}{lcccccccc}
\toprule
& \multicolumn{4}{c}{Prediction horizon (minutes)} \\ & 0 & 15 & 30 & 60  \\
\midrule
Proposed         &             \textbf{0.47} &               \textbf{0.54} &               0.57 &              0.60 \\
ANN              &              0.35 &               0.50 &               \textbf{0.57} &               \textbf{0.68} \\
SVR              &            -0.01 &               0.44 &              0.44 &              0.64 \\
k-NN &             0.05 &              0.24 &              0.36 &              0.49 \\
\bottomrule
\end{tabular}}

\end{table*}
\section{Conclusions}
In this work, we propose a dynamic linear model with time-varying coefficients to predict travel time.
The time-varying coefficients allow the linear model to represent non-linear traffic behaviors. The transition matrix consisting of these coefficients is estimated as a least-squares solution, which can be solved analytically and thus computationally efficient. The travel time predictor based on the proposed model outperforms other predictors for short-term prediction regardless of traffic situations. This can be useful to many applications, such as car navigation systems and traffic management.

\printbibliography

@inproceedings{duan2016travel,
  title={Travel time prediction with LSTM neural network},
  author={Duan, Yanjie and Lv, Yisheng and Wang, Fei-Yue},
  booktitle={2016 IEEE 19th International Conference on Intelligent Transportation Systems (ITSC)},
  pages={1053--1058},
  year={2016},
  organization={IEEE}
}

@article{yang2009reliability,
  title={The reliability of travel time forecasting},
  author={Yang, Menglong and Liu, Yiguang and You, Zhisheng},
  journal={IEEE Transactions on Intelligent Transportation Systems},
  volume={11},
  number={1},
  pages={162--171},
  year={2009},
  publisher={IEEE}
}

@article{achar2019bus,
  title={Bus Arrival Time Prediction: A Spatial Kalman Filter Approach},
  author={Achar, Avinash and Bharathi, Dhivya and Kumar, Bachu Anil and Vanajakshi, Lelitha},
  journal={IEEE Transactions on Intelligent Transportation Systems},
  year={2019},
  publisher={IEEE}
}

@inproceedings{liu2017short,
  title={Short-term travel time prediction by deep learning: A comparison of different LSTM-DNN models},
  author={Liu, Yangdong and Wang, Yizhe and Yang, Xiaoguang and Zhang, Linan},
  booktitle={2017 IEEE 20th International Conference on Intelligent Transportation Systems (ITSC)},
  pages={1--8},
  year={2017},
  organization={IEEE}
}

@article{robinson2005modeling,
  title={Modeling urban link travel time with inductive loop detector data by using the k-NN method},
  author={Robinson, Steve and Polak, John W},
  journal={Transportation research record},
  volume={1935},
  number={1},
  pages={47--56},
  year={2005},
  publisher={SAGE Publications Sage CA: Los Angeles, CA}
}

@article{salamanis2015managing,
  title={Managing spatial graph dependencies in large volumes of traffic data for travel-time prediction},
  author={Salamanis, Athanasios and Kehagias, Dionysios D and Filelis-Papadopoulos, Christos K and Tzovaras, Dimitrios and Gravvanis, George A},
  journal={IEEE Transactions on Intelligent Transportation Systems},
  volume={17},
  number={6},
  pages={1678--1687},
  year={2015},
  publisher={IEEE}
}

@article{zhang2014component,
  title={Component GARCH models to account for seasonal patterns and uncertainties in travel-time prediction},
  author={Zhang, Yanru and Haghani, Ali and Zeng, Xiaosi},
  journal={IEEE Transactions on Intelligent Transportation Systems},
  volume={16},
  number={2},
  pages={719--729},
  year={2014},
  publisher={IEEE}
}

@inproceedings{wan2014prediction,
  title={Prediction on travel-time distribution for freeways using online expectation maximization algorithm},
  author={Wan, Nianfeng and Gomes, Gabriel and Vahidi, Ardalan and Horowitz, Roberto},
  booktitle={Transportation Research Board 93rd Annual Meeting},
  number={14-3221},
  year={2014}
}

@article{ben2001network,
  title={Network state estimation and prediction for real-time traffic management},
  author={Ben-Akiva, Moshe and Bierlaire, Michel and Burton, Didier and Koutsopoulos, Haris N and Mishalani, Rabi},
  journal={Networks and spatial economics},
  volume={1},
  number={3-4},
  pages={293--318},
  year={2001},
  publisher={Springer}
}

@InProceedings{skabardonis2005real,
      title = {Real-time estimation of travel times on signalized  arterials},
      author = {Skabardonis, A. and Geroliminis, N.},
      booktitle = {International Symposium on Transportation and Traffic Theory (ISTTT)},
      pages = {387-406},
      year = {2005},
      publisher={Elsevier},
}

@inproceedings{takaba1991estimation,
  title={Estimation and measurement of travel time by vehicle detectors and license plate readers},
  author={Takaba, Sadao and Morita, Takeshi and Hada, Takashi and Usami, Tsutomu and Yamaguchi, Morie},
  booktitle={Vehicle Navigation and Information Systems Conference, 1991},
  volume={2},
  pages={257--267},
  year={1991},
  organization={IEEE}
}

@InProceedings{10.1007/978-3-030-05755-8_7,
author="Bai, Mengting
and Lin, Yangxin
and Ma, Meng
and Wang, Ping",
editor="Qiu, Meikang",
title="Travel-Time Prediction Methods: A Review",
booktitle="Smart Computing and Communication",
year="2018",
publisher="Springer International Publishing",
address="Cham",
pages="67--77",
isbn="978-3-030-05755-8"
}

@inproceedings{gao2016travel,
  title={Travel time prediction with immune genetic algorithm and support vector regression},
  author={Gao, Pan and Hu, Jianming and Zhou, Hao and Zhang, Yi},
  booktitle={2016 12th World Congress on Intelligent Control and Automation (WCICA)},
  pages={987--992},
  year={2016},
  organization={IEEE}
}

@article{castro2009online,
  title={Online-SVR for short-term traffic flow prediction under typical and atypical traffic conditions},
  author={Castro-Neto, Manoel and Jeong, Young-Seon and Jeong, Myong-Kee and Han, Lee D},
  journal={Expert systems with applications},
  volume={36},
  number={3},
  pages={6164--6173},
  year={2009},
  publisher={Elsevier}
}

@article{xia2011multistep,
  title={A multistep corridor travel-time prediction method using presence-type vehicle detector data},
  author={Xia, Jingxin and Chen, Mei and Huang, Wei},
  journal={Journal of Intelligent Transportation Systems},
  volume={15},
  number={2},
  pages={104--113},
  year={2011},
  publisher={Taylor \& Francis}
}

@inproceedings{hamner2010predicting,
  title={Predicting travel times with context-dependent random forests by modeling local and aggregate traffic flow},
  author={Hamner, Benjamin},
  booktitle={2010 IEEE International Conference on Data Mining Workshops},
  pages={1357--1359},
  year={2010},
  organization={IEEE}
}

@article{zhang2015gradient,
  title={A gradient boosting method to improve travel time prediction},
  author={Zhang, Yanru and Haghani, Ali},
  journal={Transportation Research Part C: Emerging Technologies},
  volume={58},
  pages={308--324},
  year={2015},
  publisher={Elsevier}
}

@article{fei2011bayesian,
  title={A bayesian dynamic linear model approach for real-time short-term freeway travel time prediction},
  author={Fei, Xiang and Lu, Chung-Cheng and Liu, Ke},
  journal={Transportation Research Part C: Emerging Technologies},
  volume={19},
  number={6},
  pages={1306--1318},
  year={2011},
  publisher={Elsevier}
}

@article{li2014data,
  title={A data mining based approach for travel time prediction in freeway with non-recurrent congestion},
  author={Li, Chi-Sen and Chen, Mu-Chen},
  journal={Neurocomputing},
  volume={133},
  pages={74--83},
  year={2014},
  publisher={Elsevier}
}

@article{petersen2019multi,
  title={Multi-output bus travel time prediction with convolutional LSTM neural network},
  author={Petersen, Niklas Christoffer and Rodrigues, Filipe and Pereira, Francisco Camara},
  journal={Expert Systems with Applications},
  volume={120},
  pages={426--435},
  year={2019},
  publisher={Elsevier}
}

@book{kailath2000linear,
  title={Linear estimation},
  author={Kailath, Thomas and Hassidi, Babak and Sayed, Ali H},
  year={2000},
  publisher={Prentice-Hall}
}

@article{van2009bayesian,
  title={Bayesian committee of neural networks to predict travel times with confidence intervals},
  author={van Hinsbergen, CP IJ and Van Lint, JWC and Van Zuylen, HJ},
  journal={Transportation Research Part C: Emerging Technologies},
  volume={17},
  number={5},
  pages={498--509},
  year={2009},
  publisher={Elsevier}
}

@book{feller2008introduction,
  title={An introduction to probability theory and its applications},
  author={Feller, Willliam},
  volume={2},
  year={2008},
  publisher={John Wiley \& Sons}
}

@article{hou2018network,
  title={Network scale travel time prediction using deep learning},
  author={Hou, Yi and Edara, Praveen},
  journal={Transportation Research Record},
  volume={2672},
  number={45},
  pages={115--123},
  year={2018},
  publisher={SAGE Publications Sage CA: Los Angeles, CA}
}

@article{van2008online,
  title={Online learning solutions for freeway travel time prediction},
  author={Van Lint, JWC},
  journal={IEEE Transactions on Intelligent Transportation Systems},
  volume={9},
  number={1},
  pages={38--47},
  year={2008},
  publisher={IEEE}
}

@article{kuchipudi2003development,
  title={Development of a hybrid model for dynamic travel-time prediction},
  author={Kuchipudi, Chandra and Chien, Steven},
  journal={Transportation Research Record: Journal of the Transportation Research Board},
  number={1855},
  pages={22--31},
  year={2003},
  publisher={Transportation Research Board of the National Academies}
}

@inproceedings{rice2001simple,
  title={A simple and effective method for predicting travel times on freeways},
  author={Rice, John and Van Zwet, Erik},
  booktitle={Intelligent Transportation Systems, 2001. Proceedings. 2001 IEEE},
  pages={227--232},
  year={2001},
  organization={IEEE}
}

@article{liu2006predicting,
  title={Predicting urban arterial travel time with state-space neural networks and Kalman filters},
  author={Liu, Hao and Van Zuylen, Henk and Van Lint, Hans and Salomons, Maria},
  journal={Transportation Research Record},
  volume={1968},
  number={1},
  pages={99--108},
  year={2006},
  publisher={SAGE Publications Sage CA: Los Angeles, CA}
}

@inproceedings{wu2003travel,
  title={Travel time prediction with support vector regression},
  author={Wu, Chun-Hsin and Wei, Chia-Chen and Su, Da-Chun and Chang, Ming-Hua and Ho, Jan-Ming},
  booktitle={Intelligent Transportation Systems, 2003. Proceedings. 2003 IEEE},
  volume={2},
  pages={1438--1442},
  year={2003},
  organization={IEEE}
}

@article{dia2001object,
  title={An object-oriented neural network approach to short-term traffic forecasting},
  author={Dia, Hussein},
  journal={European Journal of Operational Research},
  volume={131},
  number={2},
  pages={253--261},
  year={2001},
  publisher={Elsevier}
}

@article{van2005accurate,
  title={Accurate freeway travel time prediction with state-space neural networks under missing data},
  author={Van Lint, JWC and Hoogendoorn, SP and van Zuylen, Henk J},
  journal={Transportation Research Part C: Emerging Technologies},
  volume={13},
  number={5-6},
  pages={347--369},
  year={2005},
  publisher={Elsevier}
}

@article{dharia2003neural,
  title={Neural network model for rapid forecasting of freeway link travel time},
  author={Dharia, Abhijit and Adeli, Hojjat},
  journal={Engineering Applications of Artificial Intelligence},
  volume={16},
  number={7-8},
  pages={607--613},
  year={2003},
  publisher={Elsevier}
}

@article{innamaa2005short,
  title={Short-term prediction of travel time using neural networks on an interurban highway},
  author={Innamaa, Satu},
  journal={Transportation},
  volume={32},
  number={6},
  pages={649--669},
  year={2005},
  publisher={Springer}
}

@article{nanthawichit2003application,
  title={Application of probe-vehicle data for real-time traffic-state estimation and short-term travel-time prediction on a freeway},
  author={Nanthawichit, Chumchoke and Nakatsuji, Takashi and Suzuki, Hironori},
  journal={Transportation Research Record: Journal of the Transportation Research Board},
  number={1855},
  pages={49--59},
  year={2003},
  publisher={Transportation Research Board of the National Academies}
}

@book{boyd2004convex,
  title={Convex optimization},
  author={Boyd, Stephen and Vandenberghe, Lieven},
  year={2004},
  publisher={Cambridge university press}
}

@article{Chien:2003gg,
author = {Chien, Steven I-Jy and Kuchipudi, Chandra Mouly},
title = {{Dynamic Travel Time Prediction with Real-Time and Historic Data}},
journal = {Journal of Transportation Engineering},
year = {2003},
volume = {129},
number = {6},
pages = {608--616},
month = nov
}

@article{Park:1999ik,
author = {Park, Dongjoo and Rilett, Laurence R},
title = {{Forecasting Freeway Link Travel Times with a Multilayer Feedforward Neural Network}},
journal = {Computer-Aided Civil and Infrastructure Engineering},
year = {1999},
volume = {14},
number = {5},
pages = {357--367},
month = sep,
publisher = {Blackwell Publishers Inc.}
}

@article{Yildirimoglu:2013ip,
author = {Yildirimoglu, Mehmet and Geroliminis, Nikolas},
title = {{Experienced travel time prediction for congested freeways}},
journal = {Transportation Research Part B: Methodological},
year = {2013},
volume = {53},
pages = {45--63},
month = jul
}

@book{golub2012matrix,
  title={Matrix computations},
  author={Golub, Gene H and Van Loan, Charles F},
  volume={3},
  year={2012},
  publisher={JHU Press}
}

@article{Zhang:2003bn,
  title={Short-term travel time prediction},
  author={Zhang, Xiaoyan and Rice, John A},
  journal={Transportation Research Part C: Emerging Technologies},
  volume={11},
  number={3-4},
  pages={187--210},
  year={2003},
  publisher={Elsevier}
}



\appendices
\section{Mathematical derivations}
\subsection{Regularized least squares solution}
\label{app:RLSsolution}
A derivative of a scalar function is:
\begin{equation}
df\left( {x,y} \right) = \frac{{\partial f}}{{\partial x}}dx + \frac{{\partial f}}{{\partial y}}dy.
\end{equation}
This can be extended to a matrix form as follows:
\begin{align}
df\left( {H} \right) &= df\left( {{H_{1,1}},{H_{2,1}}, \cdots {H_{n,m}}} \right)\\
&= \frac{{\partial f}}{{\partial {H_{1,1}}}}d{H_{1,1}} + \frac{{\partial f}}{{\partial {H_{2,1}}}}d{H_{2,1}} +  \cdots \frac{{\partial f}}{{\partial {H_{n,m}}}}d{H_{n,m}}\\
&= ve{c^\top}\left( {d{H}} \right) \cdot vec\left( {\frac{{df}}{{d{H}}}} \right)\\
&= \text{tr} \left( {{{ d{H}^\top{\frac{{df}}{{d{H}}}} }} } \right),\label{eqn:traceformOfDerivative}
\end{align}
where the function $vec\left(  \cdot  \right)$ vectorizes a matrix by concatenating its columns and 
\begin{equation}
\frac{{df}}{{d{H}}} = 
\begin{bmatrix}
{\frac{{\partial f}}{{\partial {H_{1,1}}}}}&{\frac{{\partial f}}{{\partial {H_{1,2}}}}}& \cdots &{\frac{{\partial f}}{{\partial {H_{1,m}}}}}\\
{\frac{{\partial f}}{{\partial {H_{2,1}}}}}&{\frac{{\partial f}}{{\partial {H_{2,2}}}}}& \cdots &{\frac{{\partial f}}{{\partial {H_{2,m}}}}}\\
 \vdots & \vdots & \ddots & \vdots \\
{\frac{{\partial f}}{{\partial {H_{n,1}}}}}&{\frac{{\partial f}}{{\partial {H_{n,2}}}}}& \cdots &{\frac{{\partial f}}{{\partial {H_{n,m}}}}}
\end{bmatrix},
\end{equation}
\begin{equation}
d{H} = 
\begin{bmatrix}
{d{H_{1,1}}}&{d{H_{1,2}}}& \cdots &{d{H_{1,m}}}\\
{d{H_{2,1}}}&{d{H_{2,2}}}& \cdots &{d{H_{2,m}}}\\
 \vdots & \vdots & \ddots & \vdots \\
{d{H_{n,1}}}&{d{H_{n,2}}}& \cdots &{d{H_{n,m}}}
\end{bmatrix},
\end{equation}
where $H_{i,j}$ denotes the $i,j$ entry of matrix $H$.

We define the cost function of Eq.~(\ref{eqn:modifiedOP}) as $f$:
\begin{equation}
f\left({{H}_{k}} \right)  \buildrel \Delta \over =  \rho {\lambda ^{\left| {\mathbb{D}} \right|}}\left\| {{{H}_{k}}} \right\|_F^2 + \left\| {\left( {{V}_{k+1}^{\mathbb{D}} - {{H}_{k}}{V}_k^{\mathbb{D}}} \right)}{{\Lambda}} _{\left| {\mathbb{D}} \right|}^{\frac{1}{2}} \right\|_F^2.
\end{equation}
Since the cost function is convex \cite{boyd2004convex}, we utilize that the derivative at global minimum is zero. Therefore, we compute:
\begin{align}
\begin{split}
&f\left( {{{H}_{k}} + d{{H}_{k}}} \right) - f\left( {{{H}_{k}}} \right)\\
&=\text{tr} \left( 2\left( {{{H}_{k}}{V}_k^{\mathbb{D}}{\Lambda _{\left| {\mathbb{D}} \right|}}{{\left( {{V}_k^{\mathbb{D}}} \right)}^\top} - {V}_{k+1}^{\mathbb{D}}{\Lambda _{\left| {\mathbb{D}} \right|}}{{\left( {{V}_k^{\mathbb{D}}} \right)}^\top}} \right)d{H}_{k}^\top \right.\\
&\qquad+ H.O.T. \Bigr)+\rho {\lambda ^{\left| {\mathbb{D}} \right|}}\text{tr} \left( {2{{H}_{k}}d{H}_{k}^\top + H.O.T.} \right).
\end{split}
\end{align}
Here, the abbreviation \textit{H.O.T.} stands for higher order terms of $d{H}_{k}$. Assuming that $d{H}_{k}$ is small enough,
\begin{align}
\begin{split}
&f\left( {{{H}_{k}} + d{{H}_{k}}} \right) - f\left( {{{H}_{k}}} \right)\\
&\qquad=df\left( {{H}_{k}} \right)\\
&\qquad= \text{tr} \left( 2\left(  {{H}_{k}}{V}_k^{\mathbb{D}}{\Lambda _{\left| {\mathbb{D}} \right|}}{{\left( {{V}_k^{\mathbb{D}}} \right)}^\top}+\rho {\lambda ^{\left| {\mathbb{D}} \right|}}{{H}_{k}} \right.\right.\\
&\qquad\qquad\qquad\qquad\qquad\qquad\left.\left.- {V}_{k+1}^{\mathbb{D}}{\Lambda _{\left| {\mathbb{D}} \right|}}{{\left( {{V}_k^{\mathbb{D}}} \right)}^\top}\right)d{H}_{k}^\top \right).
\end{split}
\end{align}
The higher order terms are ignored since they are much smaller than the first order term $d{H}_{k}$.
By Eq.~(\ref{eqn:traceformOfDerivative}), it is confirmed that:
\begin{align}\label{eqn:dfdH}
\begin{split}
\frac{df}{d{{H}_{k}}} &=2\Bigl( {{H}_{k}}{V}_k^{\mathbb{D}}{\Lambda _{\left| {\mathbb{D}} \right|}}{{\left( {{V}_k^{\mathbb{D}}} \right)}^\top}\\
&\qquad+\rho {\lambda ^{\left| {\mathbb{D}} \right|}}{{H}_{k}} - {V}_{k+1}^{\mathbb{D}}{\Lambda _{\left| {\mathbb{D}} \right|}}{{\left( {{V}_k^{\mathbb{D}}} \right)}^\top}\Bigr).
\end{split}
\end{align}
We set the derivative equal to zero to find the global minimum, then finally:
\begin{equation}
{{\bar H }^{\mathbb{D}}}_{k + 1,k} = {V}_{k+1}^{\mathbb{D}}{{{\Lambda}} _{\left|\mathbb{D}\right|}}{\left( {{V}_k^{\mathbb{D}}} \right)^\top}{\left( {{V}_{t }^{\mathbb{D}}{{{\Lambda}} _{{\left|\mathbb{D}\right|}}}{{\left( {{V}_k^{\mathbb{D}}} \right)}^\top} + \rho {\lambda ^{{\left|\mathbb{D}\right|}}}{{I}_M}} \right)^{ - 1}}.
\end{equation}

\subsection{Regularization parameter}\label{app:regul}
The data matrix ${V}_{t }^{\mathbb{D}}{{\Lambda} _{{\left|\mathbb{D}\right|}}}{{\left( {{V}_k^{\mathbb{D}}} \right)}^\top}$ in Eq.~(\ref{eqn:modifiedOPSolution}) can be decomposed as:
\begin{equation}
{V}_{t }^{\mathbb{D}}{{\Lambda} _{{\left|\mathbb{D}\right|}}}{{\left( {{V}_k^{\mathbb{D}}} \right)}^\top}={{UD}}{{{U}}^\top},
\end{equation} 
where ${{U}}{{{U}}^\top} = {{{U}}^\top}{{U}} ={{I}_M}$ and ${D}$ is a diagonal matrix since the matrix is symmetric. Then, we can rewrite the inner part of the inversion in Eq.~(\ref{eqn:modifiedOPSolution}) as follows:
 \begin{equation}\label{eqn:inversionfree}
{{V}_{t }^{\mathbb{D}}{{\Lambda} _{{\left|\mathbb{D}\right|}}}{{\left( {{V}_k^{\mathbb{D}}} \right)}^\top} + \rho {\lambda ^{{\left|\mathbb{D}\right|}}}{{I}_M}} ={{U}}\left( {{{D}} + \rho {\lambda ^{\left|\mathbb{D}\right|}}{{I}_M}} \right){{{U}}^\top}.
\end{equation} 
Equation~(\ref{eqn:inversionfree}) proves that even if the number of training data is not enough (i.e., there are some zero values in the diagonal of ${D}$), the inversion is still available since the regularization term $\left(\rho {\lambda ^{{\left|\mathbb{D}\right|}}}{{I}_M}\right)$ is added and makes the regularized diagonal matrix $\left( {{{D}} + \rho {\lambda ^{\left|\mathbb{D}\right|}}{{I}_M}} \right)$ all non-zero on the diagonal (= full rank). Therefore, the regularization term  allows the model to be reliable in the inversion process.

\subsection{Recursive update}\label{app:recursive}
We define two matrices ${{G}_{k}^{\mathbb{D}}}$ and ${{P}_{k}^{\mathbb{D}}}$ as follows:
 \begin{equation}
{{G}_{k}^{\mathbb{D}}} \buildrel \Delta \over = {V}_{k+1}^{\mathbb{D}}{\Lambda} _{{\left|\mathbb{D}\right|}}{\left( {{V}_k^{\mathbb{D}}} \right)^\top},
\end{equation} 
 \begin{equation}
{{P}_{k}^{\mathbb{D}}} \buildrel \Delta \over = {\left( {{V}_{k}^{\mathbb{D}}{{\Lambda} _{{\left|\mathbb{D}\right|}}}}{{\left( {{V}_k^{\mathbb{D}}} \right)}^\top} + \rho {\lambda ^{{\left|\mathbb{D}\right|}}{{I}_M}} \right)^{ - 1}}.
\end{equation} 
Then we rewrite Eq.~(\ref{eqn:modifiedOPSolution}) with the multiplication of these two matrices:
 \begin{equation}\label{eqn:HwithSandC}
{{\bar H }^{\mathbb{D}}}_{k} = {{G }_{k}^{\mathbb{D}}}{{P }_{k}^{\mathbb{D}}}.
\end{equation} 
The matrix ${{G }_{k}^{\mathbb{D}\cup \tilde{d}}}$ with a new day $\tilde{d}$, which does not belong to the training set $\mathbb{D}$, can be written as:
\begin{align}
{{G }_{k}^{\mathbb{D}\cup \tilde{d}}} &= {V}_{k+1}^{\mathbb{D}\cup \tilde{d}}{{\Lambda} _{{\left| {\mathbb{D}\cup \tilde{d}}\right|}}}{\left( {{V}_k^{\mathbb{D}\cup \tilde{d}}} \right)^\top}\\
&= 
\begin{bmatrix} 
{{V}_{k+1}^{\mathbb{D}}}&{{\bf{v}}_{k+1}^{\tilde d}} 
\end{bmatrix}
\begin{bmatrix} 
{\lambda {{\Lambda} _{\left| {\mathbb{D}} \right|}}}&0\\
0&1
\end{bmatrix}
\begin{bmatrix} 
{{{\left( {{V}_k^{\mathbb{D}}} \right)}^\top}}\\
{{{\left( {{\bf{v}}_k^{\tilde d}} \right)}^\top}}
\end{bmatrix}\\
&=\lambda{{V}_{k+1}^{\mathbb{D}}}{{\Lambda} _{\left| {\mathbb{D}} \right|}}{{{\left( {{V}_k^{\mathbb{D}}} \right)}^\top}}+{\bf{v}}_{k+1}^{\tilde{d}}{\left( {{\bf{v}}_k^{\tilde{d}}}\right)^\top}\\
&= \lambda{{G }_{k}^{\mathbb{D}}} + {\bf{v}}_{k+1}^{\tilde{d}}{\left( {{\bf{v}}_k^{\tilde{d}}} \right)^\top}\label{eqn:updateG}
\end{align}
and 
\begin{align}
&{{P}_{k}^{\mathbb{D} \cup \tilde d}} = {\left( {{V}_k^{\mathbb{D} \cup \tilde d}{{\Lambda} _{\left| {\mathbb{D}\cup \tilde{d}} \right|}}{{\left( {{V}_k^{\mathbb{D} \cup \tilde d}} \right)}^\top} + \rho {\lambda ^{\left| {\mathbb{D} \cup \tilde d} \right|}}{{I}_M}} \right)^{ - 1}}\\
&= \Biggl( 
\begin{bmatrix} 
{{V}_{k}^{\mathbb{D}}}&{{\bf{v}}_{k}^{\tilde d}} 
\end{bmatrix}
\begin{bmatrix} 
{\lambda {{\Lambda} _{\left| {\mathbb{D}} \right|}}}&0\\
0&1
\end{bmatrix}
\begin{bmatrix} 
{{{\left( {{V}_k^{\mathbb{D}}} \right)}^\top}}\\
{{{\left( {{\bf{v}}_k^{\tilde d}} \right)}^\top}}
\end{bmatrix}+ \rho {\lambda ^{\left| {\mathbb{D} \cup \tilde d} \right|}}{{I}_M} \Biggr)^{ - 1}\\
&=\Bigl( \lambda {\left( {{V}_{k}^{\mathbb{D}}{{\Lambda} _{{\left|\mathbb{D}\right|}}}}{{\left( {{V}_k^{\mathbb{D}}} \right)}^\top} + \rho {\lambda ^{{\left|\mathbb{D}\right|}}{{I}_M}}\right)} + {\bf{v}}_k^{\tilde d}{{\left( {{\bf{v}}_k^{\tilde d}} \right)}^\top}\Bigr)^{ - 1}\\
&= {\left( {\lambda {{\left( {{{P}_{k}^{\mathbb{D}}}} \right)}^{ - 1}} + {\bf{v}}_k^{\tilde d}{{\left( {{\bf{v}}_k^{\tilde d}} \right)}^\top}} \right)^{ - 1}}\label{eqn:mil}\\
&= {\lambda ^{ - 1}}{{P}_{k}^{\mathbb{D}}} - \frac{{{\lambda ^{ - 1}}{{P}_{k}^{\mathbb{D}}}{\bf{v}}_k^{\tilde d}{{\left( {{\bf{v}}_k^{\tilde d}} \right)}^\top}{{P}_{k}^{\mathbb{D}}}{\lambda ^{ - 1}}}}{{1 + {\lambda ^{ - 1}}{{\left( {{\bf{v}}_k^{\tilde d}} \right)}^\top}{{P}_{k}^{\mathbb{D}}}{\bf{v}}_k^{\tilde d}}}\label{eqn:updateP},
\end{align}
where the derivations (\ref{eqn:mil}) to (\ref{eqn:updateP}) are based on the matrix inversion lemma~\cite{golub2012matrix}.
Eq.~(\ref{eqn:updateG}) and~(\ref{eqn:updateP}) show the availability of updating the matrices $\left\{ {{P}_{k}^{\mathbb{D}}},{{G}_{k}^{\mathbb{D}}} \right\}$ to $\left\{ {{P}_{k}^{\mathbb{D}\cup \tilde{d}}},{{G}_{k}^{\mathbb{D} \cup \tilde{d}}} \right\}$ with new measurements $\left\{ {{\bf{v}}_k^{\tilde{d}},{\bf{v}}_{k+1}^{\tilde{d}}} \right\}$. Therefore, we can estimate the new transition matrices ${{\bar H }^{\mathbb{D}\cup\tilde{d}}}_{k}$ with the updated matrices $\left\{ {{P}_{k}^{\mathbb{D}\cup \tilde{d}}},{{G}_{k}^{\mathbb{D} \cup \tilde{d}}} \right\}$ as in Eq.~(\ref{eqn:HwithSandC}).

\subsection{Linear minimum mean square estimator}\label{app:lmmse}
A linear estimator for the velocity vector of $i$-step ahead at time $t$ is written as:
\begin{equation}\label{eqn:linearestimator}
{\bf{v}}_{k+i|k}^{\tilde d} = {{A}_0}{\bf{v}}_k^{\tilde d} + {{A}_1}{\bf{v}}_{k - 1}^{\tilde d} +  \cdots  + {{A}_{k - 1}}{\bf{v}}_1^{\tilde d}
\end{equation}
What we need to do is to find an optimal set $\left\{ {{{A}_0},{{A}_1}, \cdots ,{{A}_{k - 1}}} \right\}$ in the minimum mean square error (MMSE) sense. From Eq.~(\ref{eqn:linearestimator}) and Eq.~(\ref{eqn:derivation_predictor}), we define the prediction error as follows:
\begin{equation}
\begin{aligned}
{\bf{v}}_{k+i}^{\tilde d} &- {\bf{v}}_{k+i|k}^{\tilde d} \\
&= \left( {{\bar H}_{k+i - 1\shortleftarrow k}^{\mathbb{D}} - {{A}_0}} \right){\bf{v}}_k^{\tilde d} - \sum\limits_{j = 1}^{k - 1} {{{A}_j}{\bf{v}}_{k - j}^{\tilde d}}  + {{\bf{n}}_{k+i-1\shortleftarrow k}^{\tilde d}}
\end{aligned}
\end{equation}
Its mean square is
\begin{align}
\begin{split}
&\mathop{\rm{E}}\left[ {\left( {{\bf{v}}_{k+i}^{\tilde d} - {\bf{v}}_{k+i|k}^{\tilde d}} \right){{\left( {{\bf{v}}_{k+i}^{\tilde d} - {\bf{v}}_{k+i|k}^{\tilde d}} \right)}^\top}} \right]\\
&= \mathop{\rm{E}}\Biggl[\left[ {\left( {{\bar H}_{k+i - 1\shortleftarrow k}^{\mathbb{D}} - {{A}_0}} \right){\bf{v}}_k^{\tilde d} - \sum\limits_{j = 1}^{k - 1} {{{A}_j}{\bf{v}}_{k-j}^{\tilde d}}  + {{\bf{n}}_{k+i-1\shortleftarrow k}^{\tilde d}}} \right]\\
&\hspace{2em}{{\left[ {\left( {{\bar H}_{k+i - 1\shortleftarrow k}^{\mathbb{D}} - {{A}_0}} \right){\bf{v}}_k^{\tilde d} - \sum\limits_{j = 1}^{k - 1} {{{A}_j}{\bf{v}}_{k-j}^{\tilde d}}  + {{\bf{n}}_{k+i-1\shortleftarrow k}^{\tilde d}}} \right]}^\top} \Biggr]
\end{split}\nonumber\\
\begin{split}
&= \mathop{\rm{E}}\Biggl[ \left[ \left( {{\bar H}_{k+i - 1\shortleftarrow k}^{\mathbb{D}} - {{A}_0}} \right){\bf{v}}_k^{\tilde d} - \sum\limits_{j = 1}^{k - 1} {{A}_j}{\bf{v}}_{k-j}^{\tilde d}  \right]\\
&\hspace{4em}\left[ \left( {\bar H}_{k+i - 1\shortleftarrow k}^{\mathbb{D}} - {A}_0 \right){\bf{v}}_k^{\tilde d} - \sum\limits_{j = 1}^{k - 1} {{A}_j}{\bf{v}}_{k-j}^{\tilde d}\right]^\top \Biggr]\\
&\hspace{7em}+ \mathop{\rm{E}}\left[ {{{\bf{n}}_{k+i-1\shortleftarrow k}^{\tilde d}}\left({{\bf{n}}_{k+i-1\shortleftarrow k}^{\tilde d}}\right)^\top}\right]
\end{split}\label{eqn:mmseCost}
\end{align}
and Eq.~(\ref{eqn:mmseCost}) is minimized when
\begin{equation}
{{A}_0} = {\bar H}_{k+i - 1\shortleftarrow k}^{\mathbb{D}}
\end{equation}
and
\begin{equation}
{{A}_j} = 0\;\text{for}\;j = 1,2, \cdots, k - 1.
\end{equation}
Therefore, the linear MMSE estimator for the velocity vector of $i$-step ahead at time step $k$ is:
\begin{align}
{\bf{v}}_{k+i|k}^{\tilde d} &= {{A}_0}{\bf{v}}_k^{\tilde d}\\
&={\bar H}_{k+i - 1\shortleftarrow k}^{\mathbb{D}}{\bf{v}}_k^{\tilde d},\label{eqn:lmmse}
\end{align}
which is equivalent to Eq.~(\ref{eqn:generalizedPrediction}).

\section{Different predictors}\label{app:methods}
In this section, we explain different travel time predictors which are compared with the proposed method.

\subsection{Instantaneous travel time}
Instantaneous travel time is calculated based on the assumption that the current state does not change with time, i.e.,
\begin{equation}
    v\left(t',x\right) = v\left(t, x\right),\;\forall t'>t,\;\forall x,
\end{equation}
when the current time is $t$. The travel time $itt\left(t\right)$ is then estimated based on this velocity field with Algorithm~\ref{alg:ett}.

\subsection{$k$-Nearest neighbor}
The $k$-Nearest neighbor ($k$-NN) method estimate unknown velocity field with the $k$ most similar (or nearest) days in the training set up to a current time $t$ in terms of euclidean distance. Specifically, the velocity field for the rest of the day (after the current time) is estimated as the average of the velocities of the $k$ nearest neighbors.. We set $k=1$, meaning that we choose the most similar day in the training set for prediction.
The travel time based on the nearest neighbor method is calculated by Algorithm~\ref{alg:ett}.

\subsection{Support vector regression}
To implement a support vector regression (SVR) method, we have followed the same procedure of the previous work~\cite{wu2003travel}. However, instead of using actual travel times as an input, we put instantaneous travel times because we think that we don't know the actual travel time at the time of estimation. We used the past 5 instantaneous travel times, i.e., $itt(t-4)$, $itt(t-3)$, ... , $itt(t)$ as input variables. These input variables are scaled to have a zero mean and a unit variance. We set the target as the actual travel time with prediction horizon $h$, $a(t+h)$. Like in Ref.~\cite{wu2003travel}, the linear kernel was chosen with the parameter setting $C=1000$ and $\tau=0.1$.

\subsection{Artificial neural network}
We have designed a simple vanilla artificial neural network (ANN) for comparison using the same scaled input and target variables as in the SVR above. We set one hidden layer with 10 neurons. We used the MLPRegressor module of Scikit-learn python package and set all the parameter settings as the default setting except for the aforementioned hidden layer setting. 
\end{document}